\documentclass[final,5p,times,twocolumn,authoryear]{elsarticle}

\usepackage[utf8]{inputenc}
\usepackage{textcomp} 
\usepackage{pifont}
\usepackage{graphicx}
\usepackage{xcolor}
\usepackage{booktabs}
\usepackage{threeparttable}
\usepackage{amssymb}
\usepackage{amsmath}
\usepackage{tabularx}
\usepackage{caption}
\usepackage{adjustbox}
\usepackage{makecell}
\usepackage{stfloats}
\usepackage{url}
\usepackage{cuted}
\usepackage{caption}
\usepackage{placeins}
\usepackage{stfloats}
\usepackage[linesnumbered,ruled,vlined]{algorithm2e}
\SetKwComment{tcp}{\%~}{}
\usepackage{tabularx}
\newcolumntype{C}{>{\centering\arraybackslash}X}

\newcommand{\cmark}{\ding{51}}
\newcommand{\xmark}{\ding{55}}

\usepackage{hyperref}
\hypersetup{
    colorlinks=true,      
    linkcolor=blue,       
    citecolor=blue,      
    urlcolor=magenta     
}

\journal{Medical Image Analysis}

\begin{document}

\begin{frontmatter}

\title{SYNAPSE-Net: A Unified Framework with Lesion-Aware Hierarchical Gating for Robust Segmentation of Heterogeneous Brain Lesions}

\author[1]{\texorpdfstring{Md. Mehedi Hassan\corref{cor1}}{Md. Mehedi Hassan}}
\author[1]{Shafqat Alam}
\author[2]{Shahriar Ahmed Seam}
\author[3]{Maruf Ahmed}

\address[1]{Department of Biomedical Engineering, Bangladesh University of Engineering and Technology, Dhaka, Bangladesh}
\address[2]{Department of Computer Science and Engineering, Bangladesh University of Engineering and Technology, Dhaka, Bangladesh}
\address[3]{Department of Electrical and Electronic Engineering, Bangladesh University of Engineering and Technology, Dhaka, Bangladesh}

\cortext[cor1]{Corresponding author}
\fntext[cor1]{Email: m.hassan.bme@gmail.com}

\begin{abstract}
Automatic segmentation of diverse heterogeneous brain lesions using multi-modal MRI is a challenging problem in clinical neuroimaging, mainly because of the lack of generalizability and high prediction variance of pathology-specific deep learning models. In this work, we propose a unified and adaptive multi-stream framework called SYNAPSE-Net to perform robust multi-pathology segmentation with reduced performance variance. The framework is based on multi-stream convolutional encoders with global context modeling and a cross-modal attention fusion strategy to ensure stable and effective multi-modal feature integration. It also employs a variance-aware training strategy to enhance the robustness of the network across diverse tasks. The framework is extensively validated using three public challenge datasets: WMH MICCAI 2017, ISLES 2022, and BraTS 2020. The results show consistent improvements in boundary accuracy, delineation quality, and stability across diverse pathologies. This proposed framework achieved a high Dice similarity coefficient (DSC) of 0.831 and a low Hausdorff distance at the 95th percentile (HD95) of 3.03 on the WMH MICCAI 2017 dataset. It also achieved the lowest HD95 of 9.69 on the ISLES 2022 dataset and the highest tumor core DSC of 0.8651 on the BraTS 2020 dataset. These results validate the robustness of the proposed framework in providing a clinically relevant computer-aided solution for automated brain lesion segmentation. Source code and pretrained models are publicly available at \url{https://github.com/mubid-01/SYNAPSE-Net-pre}.
\end{abstract}


\end{frontmatter}

\begin{strip}
  \begin{minipage}{\linewidth}
    \vspace*{-4em} 
    \centering
    \bfseries\small G\hspace{0.5em}R\hspace{0.5em}A\hspace{0.5em}P\hspace{0.5em}H\hspace{0.5em}I\hspace{0.5em}C\hspace{0.5em}A\hspace{0.5em}L\hspace{2.5em}A\hspace{0.5em}B\hspace{0.5em}S\hspace{0.5em}T\hspace{0.5em}R\hspace{0.5em}A\hspace{0.5em}C\hspace{0.5em}T \\[0.1em]

    \rule{\linewidth}{1pt} \\[0.5em]
    \includegraphics[width=0.85\linewidth]{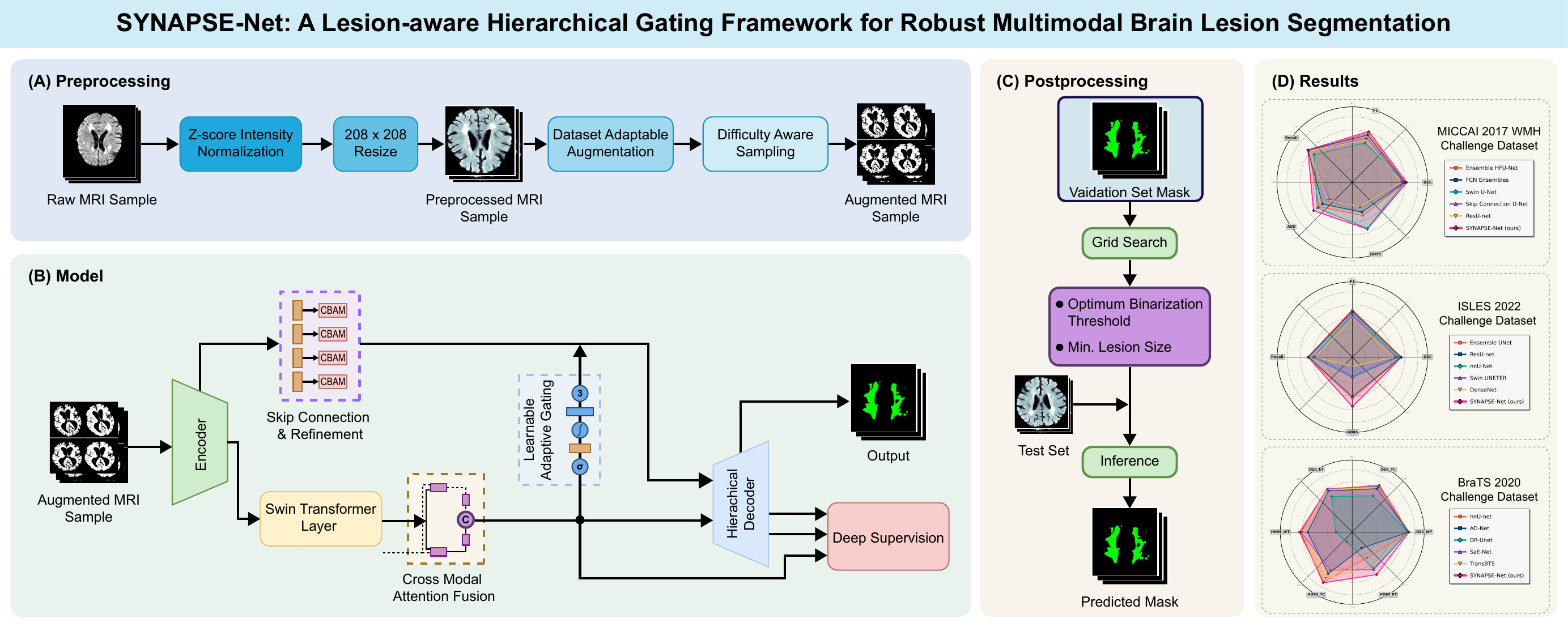}
  \end{minipage}
\end{strip}

\section{Introduction}
\label{sec:introduction}
Quantitative analysis of brain lesions from magnetic resonance imaging (MRI) is fundamental to modern neurology, underpinning the diagnosis, prognosis, and therapeutic monitoring for a spectrum of disorders, ranging from acute ischemic stroke and glioblastoma to neurodegenerative conditions such as Alzheimer’s disease. \citep{Barber1999, Capirossi2023, Grochans2022}. White matter hyperintensities (WMHs), appear as hyperintense signals on FLAIR and T2-weighted imaging sequences, are an important primary radiological sign of small vessel disease (SVD) of the brain and correlate significantly with cognitive impairment, dementia and stroke risk \citep{Du2025}. Beyond these chronic changes, glioblastoma multiforme, the most common primary brain malignancy, is becoming increasingly prevalent worldwide and substantially impacts patient mortality. This highlights the great importance of defining these tumours precisely, for surgical treatment and radiation therapy \citep{Grochans2022, Johnson2012}. Parallel to these requirements, the delineation of ischemic stroke lesions must be timely and accurate, as clinical interventions remain bound by a narrow therapeutic window\citep{Capirossi2023, Li2024, Liu2025}. Along with acute intervention, neuroimaging is vital for monitoring chronic neurological disorders. In Alzheimer’s disease, for instance, the accumulation of characteristic WMH and progressive cortical atrophy serve as essential quantitative biomarkers for longitudinal disease tracking\citep{Barber1999, Sachdev2005}. However, manual segmentation is a major bottle-neck in both clinical and research pipelines. The task is quite labour intensive and costly, which frequently restricts the possibility of performing large-scale studies. Moreover, the natural inter-observer variation is inevitable and makes these measures unreliable \citep{Sachdev2005, Vanderbecq2020}. Such limitations impede the wider application of quantitative neuroimaging, creating a pressing need for automated alternatives. Deep learning has emerged to address this, with the most recent algorithms showing that it is possible to achieve a level of performance that surpasses human raters in segmenting lesions across a broad spectrum of MRI protocols and pathologies\citep{Du2025, Duarte2024, Ronneberger2015}. 

Convolutional neural networks (CNNs) have advanced medical image analysis by automatically learning features from image datasets \citep{Ronneberger2015, Demeusy2024}. The U-Net architecture has become popular due to its encoder-decoder organization with skip connections that effectively handle contextual information while retaining spatial information to achieve accurate segmentation \citep{Ronneberger2015, Punn2022, Azad2024}. The success of the U-Net has led to specialized models to adapt for specific pathologies. U-Net variants including multi-scale extraction, attention mechanisms, and enhanced skip connections have been developed for better WMH segmentation which involves small lesions with indistinct edges \citep{Wu2019, Park2021, He2025}. U-Net and its variants were also fine-tuned for glioblastoma subregion segmentation challenges such as BraTS \citep{Walsh2022, Zheng2022, R2025}. n the context of acute ischemic stroke, the integration of ensemble CNNs, hybrid architectures, and 3D residual networks has significantly enhanced the precision of lesion identification on diffusion-weighted imaging (DWI)\citep{Tomita2020, Gheibi2023, Yassin2024}. These targeted developments have yielded effective solutions for various neurodegenerative and pathological conditions\citep{Chaki2023}.

With the growth of the domain, studies increased beyond architectural fine-tuning to methodological frameworks. One of the major challenges to address was the missing annotations, which was then solved by the paradigms of semi-supervised and weakly supervised learning that utilized unlabeled or weakly labeled data to improve performance in data-scarce settings \citep{Duarte2024, Huang2023}. At the same time, multi-center validation studies have demonstrated model robustness and generalizability across various datasets and machines \citep{Gaubert2023, Namgung2025}. Self-supervised learning approaches also improved model stability by enabling networks to learn features on unlabeled data \citep{Wu2025, ZhangX2023, Tomasetti2023}.

This methodological progress was matched by an architectural shift to overcome the inability of classic convolutional neural networks (CNNs) to capture long-range spatial interdependencies \citep{Takahashi2024, Chen2024, Chen2025}. Vision Transformers (ViTs) and Swin Transformers leverage self-attention mechanisms to model long-range spatial contexts \citep{Hatamizadeh2022, Ayoub2023, Abbaoui2024}. However, the most successful models have been hybrid architectures. For instance, hybrid frameworks such as TransUNet and Swin-Unet integrate CNNs for local feature extraction with Transformers for global context modeling, demonstrating superior performance across diverse brain pathologies\citep{He2025, Chen2024, Hatamizadeh2022, Viteri2022}. These breakthroughs were further catalyzed by multi-modal fusion strategies that effectively integrate complementary information across diverse MRI sequences, alongside the development of composite loss functions tailored for complex lesion geometries\citep{ZhangG2023, Raju2024, Cabezas2024}. 

Despite significant technological progress, two major gaps hamper the clinical translation of automated segmentation tools. The first one is the lack of generalizability. Current studies often provide task-specific solutions, with models for WMH differing from those used for glioblastoma or ischemic stroke \citep{Wu2023a, Luo2024, delaRosa2025}. This restricts clinical application, since healthcare environments require versatile models rather than single-purpose systems demanding separate maintenance \citep{Chau2025}. The second and more critical gap is variance and reliability of performances. Most of the models act unpredictably, obtaining decent dice score but performing poorly with small or non-homogenous cases, which are the circumstances that demand robust methods \citep{He2025}. This unpredictability undermines the clinical trust and hinders their application in critical cases. Addressing this performance variance, rather than incremental architectural gains, is the central focus of this work.

In order to overcome these limitations, we propose SYNAPSE-Net, a novel hybrid CNN-Transformer model that is specifically designed for effective and robust brain lesion segmentation across multimodal MRI. Unlike task-specific models, which require the design of models for specific tasks, this paper presents a universal multi-stream architecture that can be adapted to meet the needs of white matter hyperintensity segmentation, ischemic stroke segmentation, and glioblastoma segmentation. The efficacy of the proposed model is shown through a series of experiments on three publicly available datasets. The results show that the proposed model achieves state-of-the-art performance while providing improved flexibility and robustness compared to task-specific models. The main contributions of this paper are as follows:

\begin{itemize}
\item We propose a unified deep learning framework and validate its capability to achieve robust segmentation across clinically distinct heterogeneous brain lesions. We demonstrate that a fixed hybrid architecture can generalize across various tasks.

\item We introduce a new hybrid architecture designed to reduce performance variance. This is achieved by synergistically integrating multi-stream CNNs for specialized local feature extraction, Swin Transformer layers for global contextual modeling, and a Cross-Modal Attention Fusion (CMAF) mechanism for feature integration.

\item We present an adaptive variance reduction training protocol that uses data-driven difficulty-aware sampling and pathology-specific composite loss functions.

\item Our framework is rigorously evaluated on three diverse public benchmark datasets, where it outperforms numerous state-of-the-arts models. This demonstrates the superior architectural robustness and generalizability of our proposed method.
\end{itemize}

The remainder of this paper is organized as follows: Section~\ref{sec:methodology} details our proposed methodology, including the model architecture and training strategy. Section~\ref{sec:experimental_results} presents the comprehensive experimental setup and validates the utility of different components of our framework through ablation studies and results across the three target pathologies. Section~\ref{sec:limitations_and_future_works} addresses limitations and outlines future directions, and Section~\ref{sec:conclusion} provides a discussion of the results followed by concluding remarks.

\begin{figure*}[!t]
    \centering
    \includegraphics[width=\linewidth]{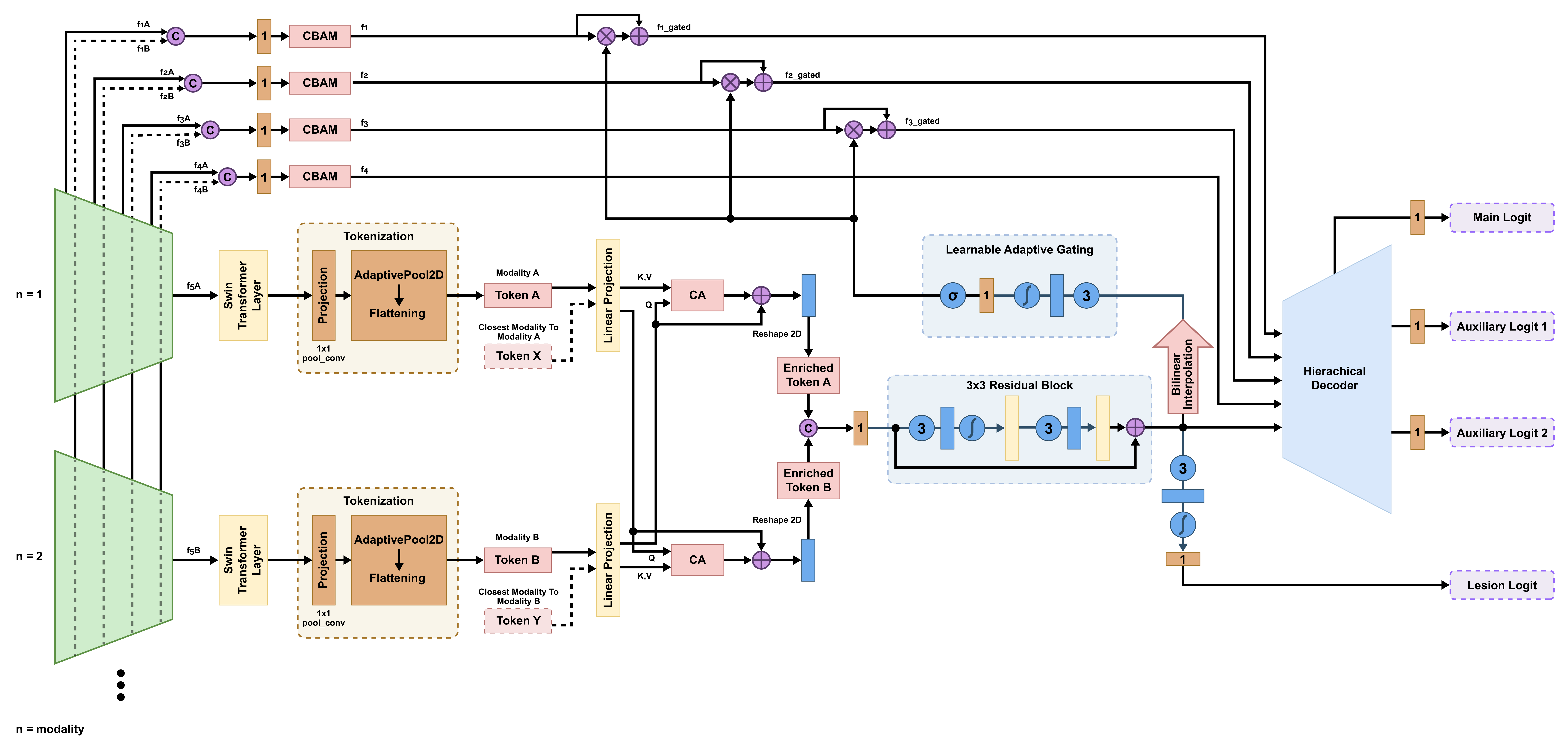}
    \caption{The SYNAPSE-Net architecture, consisting of N parallel CNN encoders, a hybrid bottleneck that fuses modalities using Swin Transformers and Cross-Modal Attention, and a Hierarchical Gated Decoder with a UNet++ backbone to generate the final segmentation.}
    \label{fig:main_architecture}
\end{figure*}

\section{Methodology}
\label{sec:methodology}

\subsection{Framework Overview}
\label{subsec:framework_overview}

SYNAPSE-Net is designed as a modular, multi-stream architecture that adaptively accommodates varying numbers of MRI modalities depending on the clinical task. As illustrated in Fig.~\ref{fig:main_architecture}, the overall architecture can be divided into three primary functional components: (i) a multi-stream CNN encoding block for modality-specific feature learning, (ii) a hybrid bottleneck block with Swin Transformers and CMAF for feature integration, and (iii) a hierarchical gated decoder with a UNet++ backbone for mask estimation. The following subsections detail the implementation of each component.

\subsection{Multi-Stream Feature Encoding and Skip Connection Refinement}
\label{subsec:encoder_and_early_fusion}

The first stage of SYNAPSE-Net learns hierarchical features from each of the N input MRI modalities. To preserve modality-specific information, each sequence is processed through an independent, architecturally identical CNN encoder before any feature fusion occurs. This approach preserves modality-specific features, preventing early fusion and potential loss of subtle yet decisive pathological signals

\begin{figure}[htbp]
    \centering
    \includegraphics[width=\linewidth]{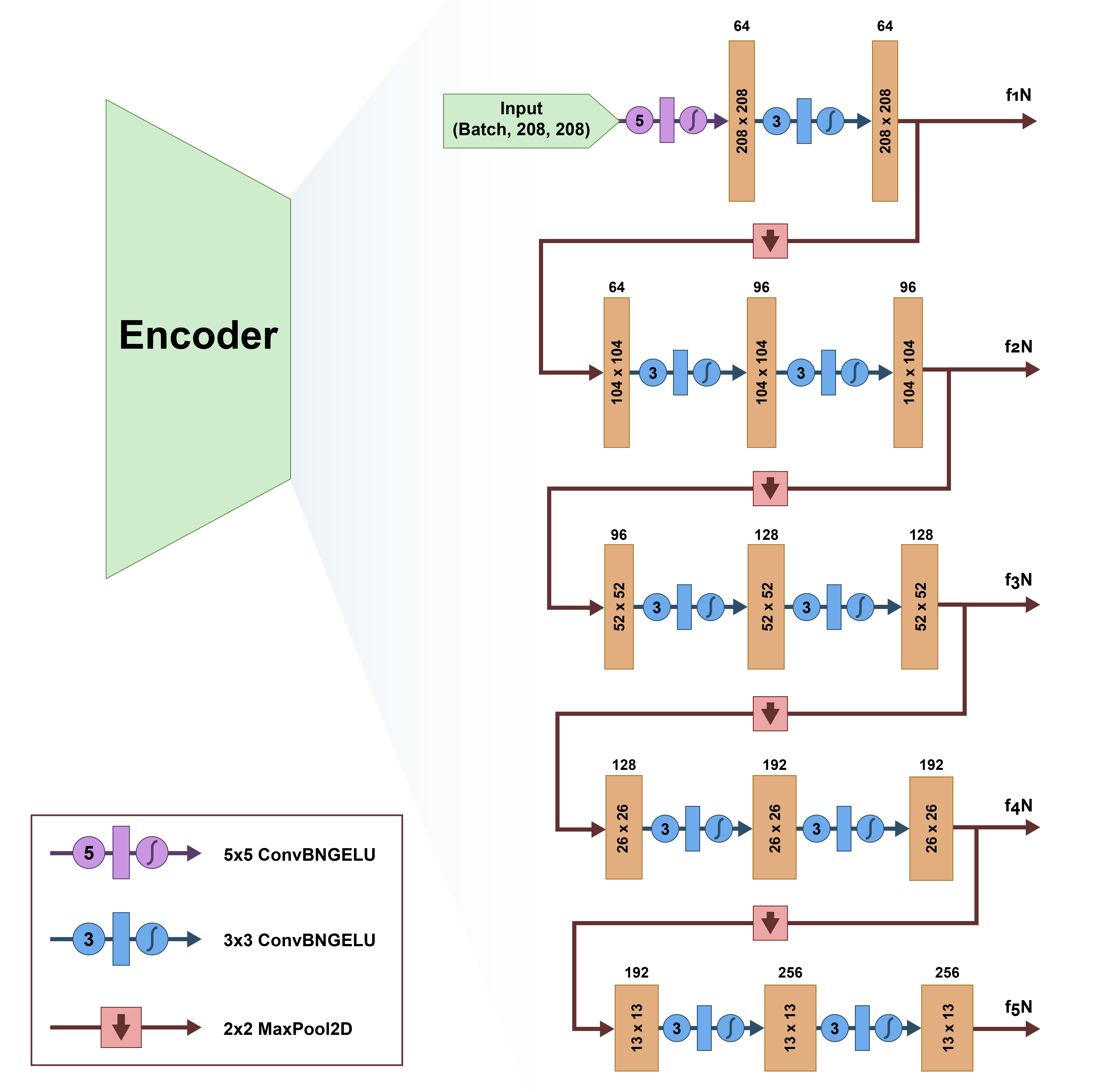}
    \caption{The five-stage Encoder for extracting five levels of feature maps ($f_1$ to $f_5$) for each input modality.}
    \label{fig:encoder_detail}
\end{figure}

Each encoder instantiates a five-stage network designed for incremental feature abstraction as depicted in Fig.~\ref{fig:encoder_detail}. The methodology systematically reduces spatial resolution by max-pooling as it concurrently increases channel depth to extract complex and semantically abstract features. This approach yields two distinct sets of outputs per modality stream: (1) a set of multi-scale feature maps, $\left[ \mathbf{f}_{i,m} | i \in \{1, \dots, 4\} \right]$, which serve as high-resolution skip connections for the decoder, and (2) a final stage deep feature map, $\mathbf{f}_{5,m}$, that captures the most abstract semantic context and serves as the input to the hybrid bottleneck. To provide the decoder with a unified input, the modality-specific skip connections from all N encoder streams are adaptively refined and fused at each resolution level, $i$. The process involves concatenating the feature maps across all streams, $\mathbf{f}_{i,\text{cat}} = [\mathbf{f}_{i,1}; \dots; \mathbf{f}_{i,N}]$. A subsequent $1 \times 1$ convolution, with learnable weights $\mathbf{W}_{\text{proj}}$, then projects this tensor back to the original channel dimension, $c_i$. This projection, which acts as a learnable channel-wise weighting of the modalities, is formulated in (\ref{eq:projection}):
\begin{equation}
    \mathbf{f}_{i,\text{proj}} = \mathbf{W}_{\text{proj}} * \mathbf{f}_{i,\text{cat}}
    \label{eq:projection}
\end{equation}
where $(*)$ denotes the convolution operation.

The projected feature map is then refined using a Convolutional Block Attention Module (CBAM) \citep{Woo2018}, which sequentially applies attention along the channel and spatial axes to refine adaptive features. The entire process that is applied on the intermediate feature map $\mathbf{F}$ can be formulated in (\ref{eq:feqn}) and (\ref{eq:fpeqn}):
\begin{align}
    \mathbf{F}' &= \mathbf{M}_c(\mathbf{F}) \otimes \mathbf{F} \label{eq:feqn}\\
    \mathbf{F}'' &= \mathbf{M}_s(\mathbf{F}') \otimes \mathbf{F}' \label{eq:fpeqn}
\end{align}
where $\otimes$ is element-wise multiplication, and $\mathbf{M}_c$ and $\mathbf{M}_s$ are the channel and spatial attention maps, respectively. The channel attention module first aggregates spatial information by using both average pooling and max pooling layers that feed into a common multi-layer perceptron (MLP) for calculating the channel attention map $\mathbf{M}_c$, as defined in (\ref{eq:channel_attention_cbam}):
\begin{equation}
    \mathbf{M}_c(\mathbf{F}) = \sigma(\text{MLP}(\text{AvgPool}(\mathbf{F})) + \text{MLP}(\text{MaxPool}(\mathbf{F})))
    \label{eq:channel_attention_cbam}
\end{equation}
The spatial attention module pools channel information into two 2D maps that get concatenated and convolved by a layer of $7\times7$ filter size in order to create the spatial attention map $\mathbf{M}_s$, which is shown in (\ref{eq:spatial_attention_cbam}):
\begin{equation}
    \mathbf{M}_s(\mathbf{F}) = \sigma(f^{7 \times 7}([\text{AvgPool}(\mathbf{F}); \text{MaxPool}(\mathbf{F})]))
    \label{eq:spatial_attention_cbam}
\end{equation}
where $\sigma$ is the sigmoid function. The final output of this stage is a set of pre-fused, attention-refined skip connections, $\{ \mathbf{f}_i | i \in \{1, 2, 3, 4\}\}$, which are passed to the hierarchical gated decoder.

\subsection{Hybrid Bottleneck: Global Context and Cross-Modal Fusion}
\label{subsec:hybrid_bottleneck}

In this stage, the deepest feature maps from each stream, $\{\mathbf{f}_{5,m}\}$, enter the hybrid bottleneck after encoding to overcome the limited receptive fields of CNNs which create semantically-rich feature representation. The process involves two primary steps: (1) Swin Transformers \citep{Liu2021} are applied for intra-modality refinement to capture global context, and (2) a fusion block employs cross-modal attention to efficiently merge the features from all streams, producing a unified bottleneck stream.

\subsubsection{Intra-Modal Refinement with Swin Transformers}
To enrich the local features extracted by the CNN with deep, global spatial context, each deep feature map, $\mathbf{f}_{5,m}$, is independently processed by dedicated Swin transformer layers before any fusion occurs. To achieve this with high computational efficiency while enabling global communication, the self-attention mechanism is applied sequentially using a shifted windowing scheme. The computation for a consecutive pair of swin transformer blocks for a layer $l$, starting with an input $\mathbf{z}^{l-1}$, is therefore formulated by (\ref{eq:swin_wmsa_full})--(\ref{eq:swin_mlp2_full}):
\begin{align}
    \mathbf{\hat{z}}^l &= \text{W-MSA}(\text{LN}(\mathbf{z}^{l-1})) + \mathbf{z}^{l-1} \label{eq:swin_wmsa_full} \\
    \mathbf{z}^l &= \text{MLP}(\text{LN}(\mathbf{\hat{z}}^l)) + \mathbf{\hat{z}}^l \label{eq:swin_mlp1_full} \\
    \mathbf{\hat{z}}^{l+1} &= \text{SW-MSA}(\text{LN}(\mathbf{z}^l)) + \mathbf{z}^l \label{eq:swin_swmsa_full} \\
    \mathbf{z}^{l+1} &= \text{MLP}(\text{LN}(\mathbf{\hat{z}}^{l+1})) + \mathbf{\hat{z}}^{l+1} \label{eq:swin_mlp2_full}
\end{align}
where W-MSA and SW-MSA denote Windowed and Shifted-Window Multi-Head Self Attention, respectively. The core self-attention computation within each MSA module is enhanced by the inclusion of a learnable relative position bias, B, in the similarity calculation. The final attention mechanism is therefore defined in (\ref{eq:self_attention_final}):
\begin{equation}
    \text{Attention}(\mathbf{Q}, \mathbf{K}, \mathbf{V}) = \text{Softmax}\left(\frac{\mathbf{Q}\mathbf{K}^T}{\sqrt{d_k}} +{B}\right)\mathbf{V}
    \label{eq:self_attention_final}
\end{equation}
where $\mathbf{Q}$, $\mathbf{K}$, $\mathbf{V}$ are the query, key, and value matrices. The output of this layer is a set of \textit{feature maps} that have been refined by both local and global self-attention, providing a rich contextual foundation for the subsequent fusion stage.

\subsubsection{Cross-Modal Attention Fusion (CMAF)}
The central module for context-enriched features fusion is the CMAF bottleneck \citep{Sun2024}. This module is designed to enable the different modality streams to query and enrich one another, creating a deeply integrated, multi-modal representation. The fusion process is adaptive based on the number of input modalities. The process begins by tokenizing each of the N context-enriched \textit{feature maps}. This is achieved by first projecting each \textit{feature map} to a new token dimension, $d_{\text{tok}}$, via a $1 \times 1$ convolution, then resizing it to a fixed spatial grid using adaptive average pooling, and finally flattening it into a sequence of tokens, $\{\mathbf{T}_{m} | m \in \{1, \dots, N\}\}$. Following tokenization, a simple and efficient pairing heuristic based on MRI physics contrast principles is then employed to group these token sequences into two final vectors for the subsequent fusion step. For even-numbered configurations, modalities are grouped by contrast affinity, such as aligning T1-weighted sequences with contrast-enhanced T1 or pairing Diffusion-Weighted Imaging with ADC maps, to exploit shared anatomical and functional cues. Conversely, for an odd number of modalities, the unpaired token sequence is concatenated to both of the primary paired vectors.

The resulting vectors $\mathbf{T}_A$ and $\mathbf{T}_B$ are processed using bi-directional multi-head cross-attention \citep{Ates2023}. As formulated in (\ref{eq:cross_attention_op}), the core operation is standard scaled dot-product attention:
\begin{equation}
    \text{CrossAtt}(\mathbf{Q}_A, \mathbf{K}_B, \mathbf{V}_B) = \text{Softmax}\left(\frac{\mathbf{Q}_A \mathbf{K}_B^T}{\sqrt{d_k}}\right)\mathbf{V}_B
    \label{eq:cross_attention_op}
\end{equation}
where the Query $\mathbf{Q}_A$ is derived from $\mathbf{T}_A$, and the Key $\mathbf{K}_B$ and Value $\mathbf{V}_B$ are derived from $\mathbf{T}_B$. The reciprocal operation is performed concurrently. The outputs are added back to their original input tokens via a residual connection and normalized to produce two enriched token sequences, $\mathbf{T}'_A$ and $\mathbf{T}'_B$, as shown in (\ref{eq:enrich_A_final}) and (\ref{eq:enrich_B_final}):
\begin{align}
    \mathbf{T}'_A &= \text{LN}(\mathbf{T}_A + \text{CrossAtt}(\mathbf{Q}_A, \mathbf{K}_B, \mathbf{V}_B)) \label{eq:enrich_A_final} \\
    \mathbf{T}'_B &= \text{LN}(\mathbf{T}_B + \text{CrossAtt}(\mathbf{Q}_B, \mathbf{K}_A, \mathbf{V}_A)) \label{eq:enrich_B_final}
\end{align}

Lastly, the enriched tokens get de-tokenized back into a spatial grid, concatenated, projected through a $1 \times 1$ convolution, and fed through a final residual block. The end-result of this procedure is the final bottleneck tensor, \textit{center}, which acts as the rich semantic base for the decoder.

\subsection{Hierarchical Gated Decoder}
\label{subsec:hierarchical_decoder}
\begin{figure}[htbp]
    \centering
    \includegraphics[width=\linewidth]{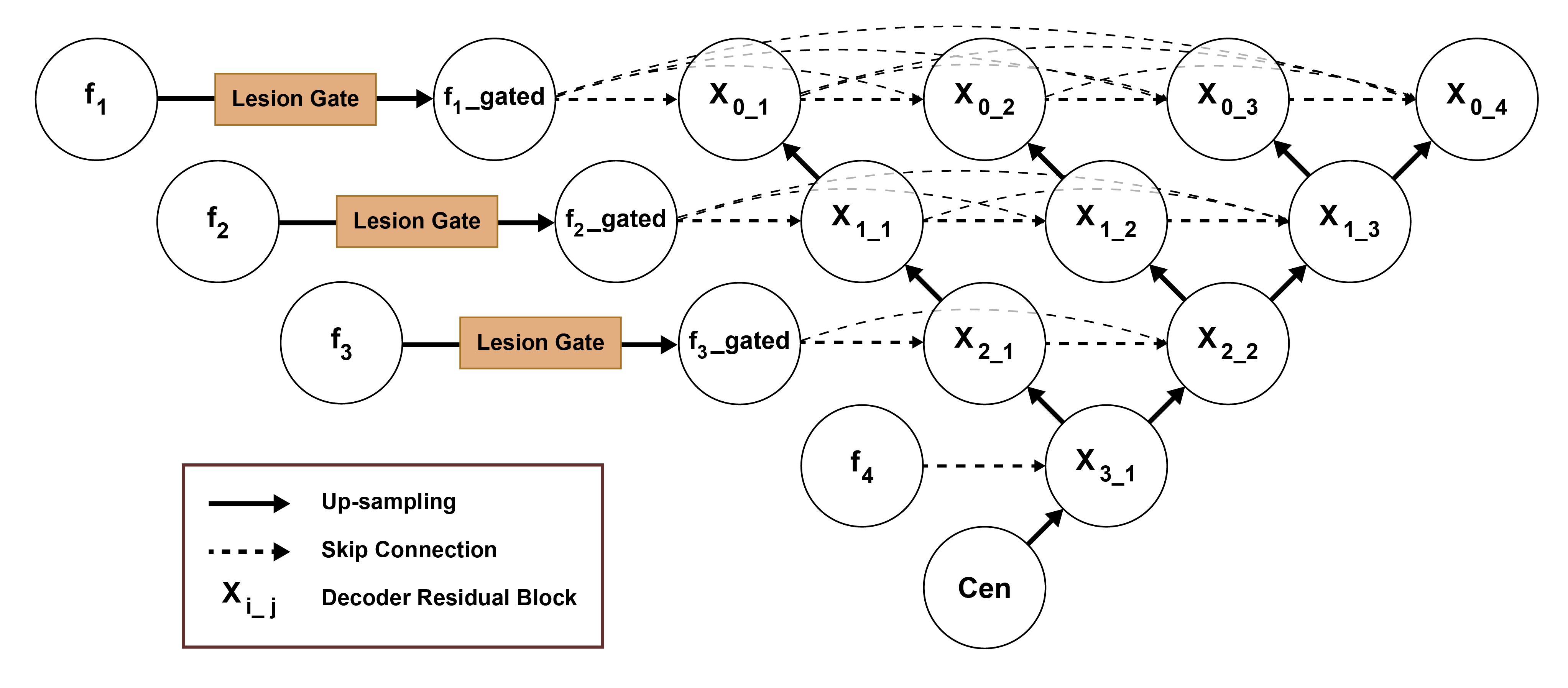}
    \caption{Conceptual diagram of the Hierarchical Gated Decoder, where Lesion Gate modules refine skip connections ($f_1$–$f_4$) before feeding them into the dense UNet++ decoder nodes ($X_{i,j}$).}
    \label{fig:decoder_conceptual}
\end{figure}

\begin{figure}[htbp]
    \centering
    \includegraphics[width=\linewidth]{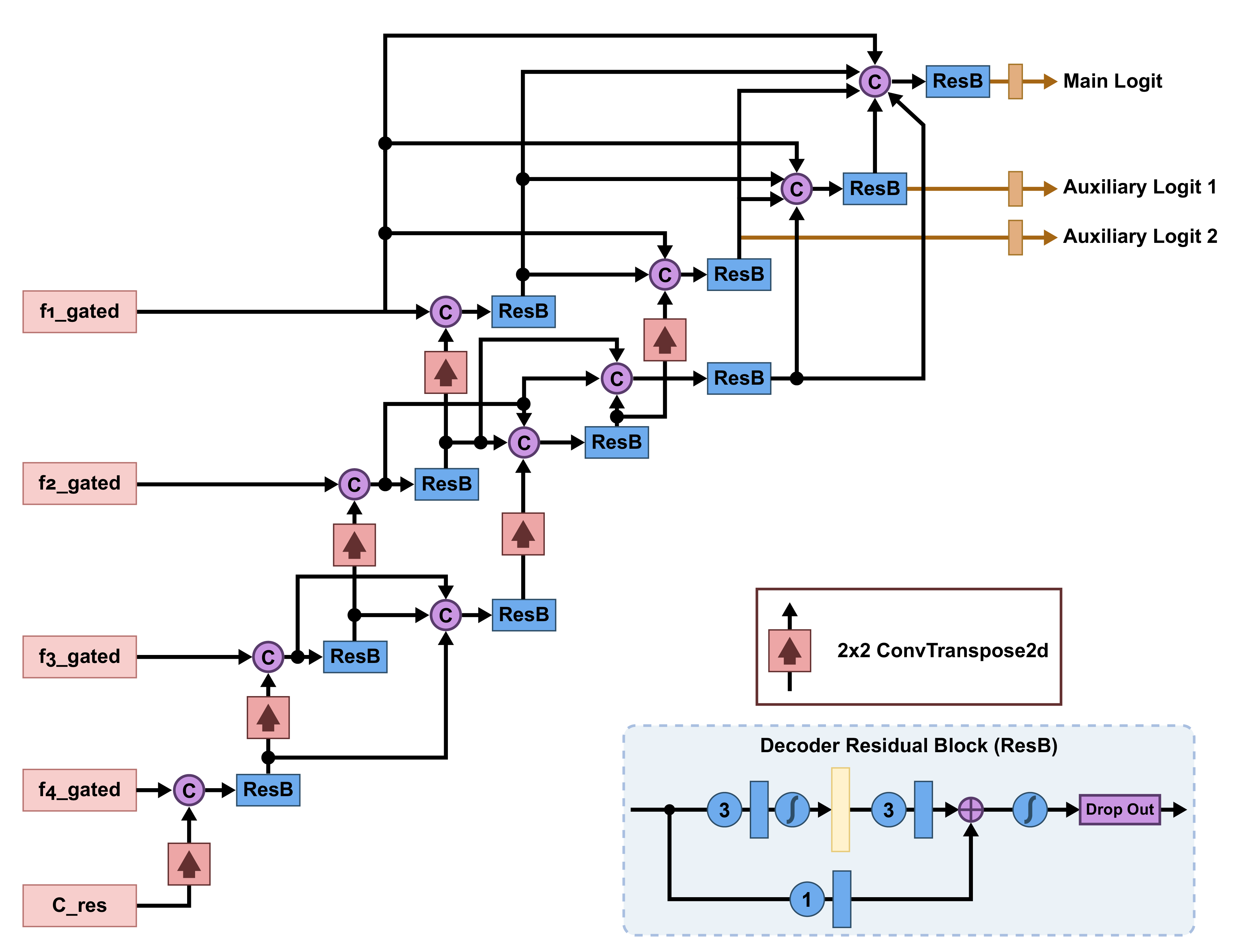}
    \caption{The hierarchical gated decoder realization using the dense UNet++ backbone, where the gated skip connections ($f_{i,\text{gated}}$) and bottleneck tensor are utilized.}
    \label{fig:decoder_block_detail}
\end{figure}

The final stage of the SYNAPSE-Net is a nested decoder designed for high-fidelity mask reconstruction. This architecture is built upon the robust foundation of the UNet++ topology and is significantly enhanced by our proposed \textit{Hierarchical Adaptive Gating mechanism}. This novel mechanism performs top-down feature modulation, using the rich pathological context derived from the network's bottleneck to create a cascade of progressively refined attention. The conceptual data flow is illustrated in Fig.~\ref{fig:decoder_conceptual}, with the practical implementation detailed in Fig.~\ref{fig:decoder_block_detail}.

\subsubsection{The Hierarchical Adaptive Gating Mechanism}
To refine the features before they reach the decoder, we incorporate a Hierarchical Gating mechanism which performs top-down feature modulation to refine the pre-fused skip connections ($\mathbf{f}_i$). This is achieved through a series of \texttt{LesionGate} modules. As formulated in (\ref{eq:gate_signal}) and (\ref{eq:gated_skip}), each module takes a skip connection and a guidance signal from a deeper, more semantically abstract layer to compute a spatially-refined, gated feature map, $\mathbf{f}_{i,\text{gated}}$.

The core of the \texttt{LesionGate} is the generation of a spatial attention gate, $\mathbf{S}_{\text{gate}}$, from the guidance signal, $\mathbf{g}$:
\begin{equation}
    \mathbf{S}_{\text{gate}} = \mathcal{C}(\mathcal{U}(\mathbf{g}))
    \label{eq:gate_signal}
\end{equation}
where $\mathcal{C}$ is a small convolutional block and $\mathcal{U}(\cdot)$ is an up-sampling operation. This gate then additively refines the original skip connection features via a residual connection, which is critical for stable training:
\begin{equation}
    \mathbf{f}_{i,\text{gated}} = \mathbf{f}_i + \left( \mathbf{f}_i \otimes \sigma(\mathbf{S}_{\text{gate}}) \right)
    \label{eq:gated_skip}
\end{equation}
where $\sigma$ is the sigmoid activation and $\otimes$ denotes element-wise multiplication.

This process is applied in a top-down cascade, creating a flow of progressively refined attention. As detailed in (\ref{eq:gate_level3})--(\ref{eq:gate_level1}), the cascade begins with the most abstract features from the bottleneck and culminates at the highest resolution.
\begin{enumerate}
    \item \textbf{Level 3 Gating:} The initial guidance signal is the final bottleneck tensor, \textit{center}. It is used to modulate the deepest skip connection, $\mathbf{f}_3$:
    \begin{equation}
        \mathbf{f}_{3,\text{gated}} = \text{LesionGate}(\mathbf{f}_3, \mathbf{\textit{center}})
        \label{eq:gate_level3}
    \end{equation}

    \item \textbf{Level 2 Gating:} The output of the first decoder stage, $\mathbf{x}_{3,1}$, now serves as a more spatially refined guidance signal to modulate the mid-level skip connection, $\mathbf{f}_2$:
    \begin{equation}
        \mathbf{f}_{2,\text{gated}} = \text{LesionGate}(\mathbf{f}_2, \mathbf{x}_{3,1})
        \label{eq:gate_level2}
    \end{equation}

    \item \textbf{Level 1 Gating:} This process culminates at the highest resolution, where the output of the second decoder stage, $\mathbf{x}_{2,2}$, serves as a highly specific guidance signal to modulate the shallowest skip connection, $\mathbf{f}_1$:
    \begin{equation}
        \mathbf{f}_{1,\text{gated}} = \text{LesionGate}(\mathbf{f}_1, \mathbf{x}_{2,2})
        \label{eq:gate_level1}
    \end{equation}
\end{enumerate}
This hierarchical cascade ensures that the attention applied at each level is progressively more informed, allowing the model to apply precise, semantically-aware attention to the fine-grained features critical for accurate boundary delineation.

\subsubsection{UNet++ Decoder Backbone}
The decoder architecture is built upon the UNet++ framework, which employs a series of nested, dense skip connections to bridge the semantic gap between encoder and decoder features, with the practical implementation detailed in Fig.~\ref{fig:decoder_block_detail}. Let $\mathbf{x}_{i,j}$ denote the output of a decoder node, where `i' indexes the down-sampling level and `j' indexes the convolutional layer along the dense skip pathway.

As formulated in (\ref{eq:unetpp_node_general}), the computation for any node is a function of the concatenated outputs of all previous nodes in the same row and the up-sampled output from the deeper, adjacent node:
\begin{equation}
    \mathbf{x}_{i,j} = \mathcal{H}\left(\left[ \{\mathbf{x}_{i,k}\}_{k=0}^{j-1}, \mathcal{U}(\mathbf{x}_{i+1,j-1}) \right]\right) \quad \text{for } j > 0
    \label{eq:unetpp_node_general}
\end{equation}
where $\mathcal{H}(\cdot)$ is the \texttt{ResidualBlock} operation, $[\cdot]$ denotes channel-wise concatenation, and $\mathcal{U}(\cdot)$ is a \texttt{ConvTranspose2d} up-sampling operation.

The inputs to the first node of each pathway ($j=1$) are the corresponding gated skip connection and the up-sampled output from the node below. The information flow begins at the deepest level with the \textit{center} tensor from the bottleneck and the $\mathbf{f}_4$ skip connection (which bypasses gating due to its inherent semantic alignment with the bottleneck features). The computation for the first node in each of the four main decoder pathways is therefore a specific application of this general rule, as shown in (\ref{eq:node_x31})--(\ref{eq:node_x01}):
\begin{align}
    \mathbf{x}_{3,1} &= \mathcal{H}([\mathbf{f}_4, \mathcal{U}(\mathbf{center})]) \label{eq:node_x31} \\
    \mathbf{x}_{2,1} &= \mathcal{H}([\mathbf{f}_{3,\text{gated}}, \mathcal{U}(\mathbf{x}_{3,1})]) \label{eq:node_x21} \\
    \mathbf{x}_{1,1} &= \mathcal{H}([\mathbf{f}_{2,\text{gated}}, \mathcal{U}(\mathbf{x}_{2,1})]) \label{eq:node_x11} \\
    \mathbf{x}_{0,1} &= \mathcal{H}([\mathbf{f}_{1,\text{gated}}, \mathcal{U}(\mathbf{x}_{1,1})]) \label{eq:node_x01}
\end{align}
This hierarchical structure, which integrates features across multiple semantic levels, allows the model to produce highly accurate and detailed segmentation masks.

\subsubsection{Deep Supervision and Final Output Generation}
To ensure robust gradient flow, the framework integrates deep supervision by generating multiple outputs. We leverage the nested UNet++ decoder by attaching 1x1 convolutional heads to each of the final three decoder nodes ($\mathbf{x}_{0,2}$, $\mathbf{x}_{0,3}$, and $\mathbf{x}_{0,4}$), which produce two auxiliary logit maps ($\mathbf{z}_{\text{aux},k}$) and the main logit map ($\mathbf{z}_{\text{main}}$). A separate auxiliary head on the bottleneck's \textit{center} tensor generates a low-resolution lesion localization map, $\mathbf{z}_{\text{lesion}}$, to directly supervise the network's most abstract features. While a weighted loss is applied to all of these outputs during training to provide intermediate gradients, only the main logit map is used for the final inference.

\subsection{Model Loss Function}
\label{subsec:loss_functions}
The training is guided by an adaptive, composite loss function, $\mathcal{L}_{\text{total}}$. It is a sum of a main loss ($\mathcal{L}_{\text{main}}$), an auxiliary deep supervision loss ($\mathcal{L}_{\text{aux}}$), and a bottleneck lesion localization loss ($\mathcal{L}_{\text{lesion}}$), weighted by $\lambda_{\text{lesion}}$ to regulate the auxiliary localization signal, ensuring it guides the network without overpowering the primary segmentation task, as formulated in (\ref{eq:total_loss}).
\begin{equation}
\mathcal{L}_{\text{total}} = \mathcal{L}_{\text{main}} + \mathcal{L}_{\text{aux}} + \lambda_{\text{lesion}} \mathcal{L}_{\text{lesion}}
\label{eq:total_loss}
\end{equation}

\subsubsection{Foundational Loss Functions}
\textit{Dice and Focal Loss:}
The dice loss maximizes overlap between predictions $P$ and ground truth $T$, as defined in (\ref{eq:dice_loss}):
\begin{equation}
\mathcal{L}_{\text{Dice}}(P, T) = 1 - \frac{2 \sum_{i=1}^N P_i T_i + \epsilon}{\sum_{i=1}^N P_i + \sum_{i=1}^N T_i + \epsilon}
\label{eq:dice_loss}
\end{equation}
where $\epsilon$ is a smoothing constant. Focal loss addresses class imbalance by scaling the cross-entropy loss by $(1-p_t)^{\gamma}$, shown in (\ref{eq:focal_loss}):
\begin{equation}
\mathcal{L}_{\text{Focal}}(p, y) = -\alpha_t (1 - p_t)^{\gamma} \log(p_t)
\label{eq:focal_loss}
\end{equation}
where $p_t$ is the probability of the correct class, and $\gamma$ and $\alpha_t$ are focusing and balancing parameters.

\textit{Tversky and Boundary Loss:}
The Tversky loss generalizes dice to control the trade-off between false positives and false negatives (FN) via parameters $\alpha$ and $\beta$, given by (\ref{eq:tversky_loss}):
\begin{multline}
\mathcal{L}_{\text{Tversky}}(P, T; \alpha, \beta) = 1 - \\
\frac{\sum_i P_i T_i + \epsilon}{\sum_i P_i T_i + \alpha \sum_i P_i (1 - T_i) + \beta \sum_i (1 - P_i) T_i + \epsilon}
\label{eq:tversky_loss}
\end{multline}
Setting $\beta > \alpha$ penalizes FNs more heavily. We merge Focal and Tversky losses into a weighted sum, balanced by scalar weights $w_f$ and $w_t$, as expressed in (\ref{eq:focal_tversky_loss}):
\begin{equation}
\mathcal{L}_{\text{FocalTversky}} = w_f\mathcal{L}_{\text{Focal}} + w_t\mathcal{L}_{\text{Tversky}}
\label{eq:focal_tversky_loss}
\end{equation}
To improve geometric accuracy, a `Boundary Loss' \citep{Kervadec2018} penalizes predictions far from the ground truth boundary using a pre-computed Euclidean distance map $d_G$, averaged over the total number of voxels, $N$, as shown in (\ref{eq:boundary_loss}):
\begin{equation}
\mathcal{L}_{\text{Boundary}}(P, d_G) = \frac{1}{N} \sum_{i=1}^N P_i \cdot (d_G)_i
\label{eq:boundary_loss}
\end{equation}

\subsubsection{Pathology-Specific Loss Configurations}
We use adapted total loss for each pathology. In our formulations, $z_{\text{main}}$, $z_{\text{aux},k}$, and $z_{\text{lesion}}$ are the raw logit outputs. The ground-truth mask is $y$, $\sigma(\cdot)$ is the sigmoid function, ($\downarrow$) is a downsampling operator, and $w_k$ weights the $k$-th auxiliary output.

\begin{table*}[!t]
\centering
\begin{threeparttable}
\caption{Pathology-Adapted Data Augmentation Protocols}
\label{tab:augmentation_protocols}
\scriptsize
\begin{tabular*}{\textwidth}{@{\extracolsep{\fill}}lccccccc@{}}
\toprule
\textbf{Dataset} & \textbf{\makecell{Random\\Flips}} & \textbf{\makecell{Affine\\Probability}} & \textbf{\makecell{Rotation\\Range}} & \textbf{\makecell{Scale\\Range}} & \textbf{\makecell{Elastic\\Deformation}} & \textbf{\makecell{Photometric\\Augmentation}} & \textbf{\makecell{Channel\\Dropout}} \\
\midrule
WMH (Highest) & 50\% & 75\% & $\pm 20^{\circ}$ & $\pm 20\%$ & 50\% & 30-50\% & 50\% \\
ISLES (Medium) & 50\% & 75\% & $\pm 15^{\circ}$ & $\pm 15\%$ & 30\% & 30\% & 25\% \\
BraTS (Lowest) & 50\% & 50\% & $\pm 10^{\circ}$ & $\pm 10\%$ & - & 15-20\% & - \\
\bottomrule
\end{tabular*}
\begin{tablenotes}
\scriptsize
\item \textit{Note}: The intensity of on-the-fly transforms was tiered from highest (WMH) to lowest (BraTS) to match the specific characteristics of each pathology and dataset size. Dash (-) indicates the augmentation was not applied.
\end{tablenotes}
\end{threeparttable}
\end{table*}

\textit{Composite Loss Function for Vascular Lesion (WMH \& ISLES):}
We use a composite loss where $\lambda_b$ modulates the boundary consistency to maximize shape fidelity, as defined in (\ref{eq:vascular_main})--(\ref{eq:vascular_lesion}):
\begin{alignat}{2}
\mathcal{L}_{\text{main}}^{\text{Vascular}} & &&= \mathcal{L}_{\text{FocalTversky}}(z_{\text{main}}, y) + \lambda_{b} \mathcal{L}_{\text{Boundary}}(\sigma(z_{\text{main}}), y) \label{eq:vascular_main}\\
\mathcal{L}_{\text{aux}}^{\text{Vascular}}  & &&= \sum_{k} w_k \cdot \mathcal{L}_{\text{FocalTversky}}(z_{\text{aux},k}, y) \label{eq:vascular_aux}\\
\mathcal{L}_{\text{lesion}}            & &&= \mathcal{L}_{\text{Focal}}(z_{\text{lesion}}, \downarrow y) \label{eq:vascular_lesion}
\end{alignat}

\textit{Composite Loss Function for Brain Tumor (BraTS):}
Here, we define the lesion target, $y_{\text{lesion}}$, as a binary mask of the entire tumor region, derived from the multi-class ground truth via an indicator function, $y_{\text{lesion}} = (\mathbf{1}_{y>0})$. The resulting loss configuration is expressed in (\ref{eq:brats_main})--(\ref{eq:brats_lesion}):
\begin{alignat}{2}
\mathcal{L}_{\text{main}}^{\text{BraTS}} & &&= \mathcal{L}_{\text{Dice}}(z_{\text{main}}, y) \label{eq:brats_main}\\
\mathcal{L}_{\text{aux}}^{\text{BraTS}}  & &&= \sum_{k} w_k \cdot \mathcal{L}_{\text{Dice}}(z_{\text{aux},k}, y) \label{eq:brats_aux}\\
\mathcal{L}_{\text{lesion}}            & &&= \mathcal{L}_{\text{Focal}}(z_{\text{lesion}}, \downarrow y_{\text{lesion}}) \label{eq:brats_lesion}
\end{alignat}

\section{Experimental Results}
\label{sec:experimental_results}

\subsection{Datasets}
\label{subsec:datasets}

We evaluated the proposed framework on three distinct, publicly available benchmark datasets\footnote{\url{https://doi.org/10.34894/AECRSD}}\textsuperscript{,}%
\footnote{\url{https://zenodo.org/records/7960856}}\textsuperscript{,}%
\footnote{\url{https://www.med.upenn.edu/cbica/brats2020/}}, each selected to represent a unique and challenging clinical scenario. 

\subsubsection{MICCAI 2017 WMH Challenge Dataset}
To address the segmentation of chronic small vessel disease, we employed the MICCAI 2017 White Matter Hyperintensities (WMH) Segmentation Challenge dataset \citep{Kuijf2022}. The training cohort consists of 60 subjects acquired across three institutes: UMC Utrecht, NUHS Singapore, and VU Amsterdam. For our experiments, we partitioned this official training set into 48 subjects for model optimization and 12 for internal validation. We utilized the provided 3D T1-weighted images and 2D multi-slice FLAIR images. The FLAIR sequences exhibit significant anisotropy with a slice thickness of 3.00 mm and variable in-plane resolutions ranging from $0.95$ mm to $1.21$ mm. The held-out test set consists of 110 subjects and introduces a significant domain shift by including data from two additional scanners not represented in the training cohort, including a 1.5T system. All quantitative results for this task reported in this study are derived from the evaluation of these 110 held-out subjects.

\subsubsection{ISLES 2022 Challenge Dataset}
To benchmark performance on acute to sub-acute ischemic stroke, we utilized the Ischemic Stroke Lesion Segmentation (ISLES) 2022 Challenge dataset \citep{HernandezPetzsche2022}. This multi-center cohort comprises 250 annotated subjects, which we partitioned into 200 for model optimization and 50 for internal validation. Although the challenge includes a held-out test set of 150 cases, access to this set was not available for independent evaluation. So, all quantitative results for this task are reported based on our validation cohort. The data was acquired across three independent centers using heterogeneous hardware, including 3T Philips systems and both 1.5T and 3T Siemens systems. For this task, we utilized the Diffusion-Weighted Imaging (DWI, $b=1000$ s/mm$^2$) and the corresponding Apparent Diffusion Coefficient (ADC) map. The dataset exhibits substantial acquisition variability: Repetition Time (TR) ranges from 3175 to 16439 ms, Echo Time (TE) from 55 to 91 ms, and slice thickness from 2.0 to 6.5 mm, with in-plane resolutions varying between $0.88 \times 0.88$ mm$^2$ and $2.0 \times 2.0$ mm$^2$.

\subsubsection{BraTS 2020 Challenge Dataset}
Finally, for the segmentation of heterogeneous brain tumors, we employed the BraTS 2020 Challenge dataset \citep{Menze2015, Bakas2017, Bakas2018}. This benchmark consists of multi-parametric MRI (mpMRI) scans from 660 diffuse glioma patients, partitioned into three cohorts: a training set (369 cases), a validation set (125 cases), and a testing set (166 cases). Each subject includes four standard modalities: native T1-weighted (T1), post-contrast T1-weighted (T1ce), T2-weighted (T2), and T2 Fluid-Attenuated Inversion Recovery (FLAIR). All imaging volumes are provided in a highly standardized state, having been skull-stripped, co-aligned to the SRI24 anatomical atlas, and resampled to a $1mm^3$ isotropic voxel resolution. The ground truth annotations comprise three sub-regions: the enhancing tumor (ET), the tumor core (TC) (which encompasses the ET and the necrotic/non-enhancing core), and the whole tumor (WT). The ground truth labels for the official validation and testing cohorts are withheld by the challenge organizers for ranking purposes.

\begin{table}[b]
\centering
\begin{threeparttable}
\caption{Network Training Parameters}
\label{tab:network_params}
\scriptsize
\begin{tabular*}{\columnwidth}{@{\extracolsep{\fill}}lccc@{}}
\toprule
& \textbf{WMH} & \textbf{ISLES} & \textbf{BraTS} \\
\midrule
\textbf{Learning Rate} & 1e-4 & 1e-4 & 5e-5 \\
\textbf{Epochs} & 150 & 120 & 300 \\
\textbf{Batch Size} & 18 & 18 & 8 \\
\textbf{Weight Decay} & 1.5e-4 & 1.5e-4 & 1e-4 \\
\textbf{Optimizer} & \multicolumn{3}{c}{AdamW} \\
\textbf{LR Scheduler\tnote{*}} & \multicolumn{3}{c}{\cmark (15-epoch warmup + cosine)} \\
\textbf{5-Fold CV} & \cmark & \cmark & \xmark \\
\textbf{Test-Time Aug.} & \cmark & \cmark & \xmark \\
\bottomrule
\end{tabular*}
\begin{tablenotes}
\scriptsize
\item[*] Linear warmup followed by cosine annealing decay.
\end{tablenotes}
\end{threeparttable}
\end{table}

\subsection{Implementation Details}
\label{subsec:implementation_details}

A standardized preprocessing pipeline was also applied for all datasets. Initial structural registration was done dataset-specifically, whereby for the WMH Challenge dataset skull stripping was done using FSL’s Brain Extraction Tool (BET) \citep{Jenkinson2012}, and for ISLES 2022, rigid co-registration for the skull-stripped ADC and DWI volumes was performed. For the BraTS 2020 dataset, the images came fully preprocessed from the challenge organizers. To ensure preprocessing consistency across all data, a Z-score intensity normalization was applied with statistics derived from the 2nd to 98th percentile of brain tissue intensities from the subjects of each respective dataset. Additionally, all axial slices were then cropped or zero-padded to a uniform 208 × 208 matrix to serve as the 2D network input. To mitigate overfitting, we employed an adaptive, on-the-fly data augmentation strategy wherein the intensity of the augmentation protocol was specifically tailored to each pathology, as detailed in Table~\ref{tab:augmentation_protocols}. In the case of vascular tasks (i.e. WMH and ISLES datasets), we enhanced robustness on small lesions by employing a difficulty-aware sampling strategy. This method oversampled 2D slices containing lesions smaller than a pre-calculated volume percentile. The hyperparameters for each training configuration, including the initial learning rate, total epochs, and optimizer details are summarized in Table~\ref{tab:network_params}. All models were implemented in PyTorch and trained on a workspace with an NVIDIA Quadro RTX 5000 16GB GPU and an Intel Xenon® W-2265 3.50 GHz processor.

\subsection{Inference Protocol and Post-processing}
\label{sec:inference_protocol}

To ensure fair and reproducible evaluation across all clinical tasks, we implemented a rigorous two-step inference protocol (Algorithm~\ref{alg:inference}) that systematically separates model inference from post-processing parameter selection. Hyperparameter tuning stage only uses the validation dataset. For each 3D validation case, we generate continuous probability volumes by means of sliding-window inference with a Gaussian weight function. This technique creates smooth 3D probability maps with no artifacts by analyzing axial slices with significant overlap and averaging the predictions in those regions.

We then perform a comprehensive grid search for the  binarization threshold $\tau \in [0.10, 0.80]$ and minimum lesion size $S_{\text{min}} \in \{2, 3, \dots, 15\}$ voxels. The optimal parameters are selected in a way so that the mean DSC across the validation cohort is maximized, which is shown in (\ref{eq:infer}):
\begin{equation}
(\tau_{\text{opt}}, S_{\text{min,opt}}) = \underset{\tau, S_{\text{min}}}{\arg\max} \; \mathbb{E}[\text{DSC}(\hat{\mathbf{Y}}_{\text{val}}, \mathbf{Y}_{\text{val}})]
\label{eq:infer}
\end{equation}

The evaluation stage uses the computed optimal parameters $(\tau_{\text{opt}}, S_{\text{min,opt}})$ to the final evaluation. Sliding window inference generates probability maps, which we binarize at $\tau_{\text{opt}}$ and refine using 3D connected-component analysis to remove all predictions smaller than $S_{\text{min,opt}}$.

\begin{algorithm}[htbp]
\caption{Inference and Hyperparameter Tuning Protocol}
\label{alg:inference}

\KwIn{
    $\mathcal{M}$ \tcp*{Trained model} 
    $\mathcal{D}_{\text{val}} = \{\mathbf{X}_{\text{val}}, \mathbf{Y}_{\text{val}}\}$ \tcp*{Validation set}
    $\mathcal{D}_{\text{test}} = \{\mathbf{X}_{\text{test}}, \mathbf{Y}_{\text{test}}\}$ \tcp*{Test set}
}
\KwOut{
    $\mathcal{R}_{\text{metrics}}$ \tcp*{Final evaluation metrics}
    $\tau_{\text{opt}}$ \tcp*{Optimal threshold}
    $S_{\text{min,opt}}$ \tcp*{Optimal minimum lesion size}
}

\tcp{Stage 1: Hyperparameter Tuning on Validation Set}
$\mathcal{P}_{\text{val}} \leftarrow F_{\text{inference}}(\mathbf{X}_{\text{val}}, \mathcal{M})$\;
$\tau_{\text{opt}} \leftarrow 0.5$; $S_{\text{min,opt}} \leftarrow 0$; $\mathcal{S}_{\text{best}} \leftarrow 0$\;
\For{$\tau \in [0.10, 0.80]$}{
    \For{$S_{\text{min}} \in \{2, 3, \dots, 15\}$}{
        $\mathcal{L}_{\text{scores}} \leftarrow \emptyset$\;
        \ForEach{$(\mathbf{P}_i, \mathbf{Y}_i) \in (\mathcal{P}_{\text{val}}, \mathbf{Y}_{\text{val}})$}{
            $\hat{\mathbf{Y}}_i \leftarrow F_{\text{postprocess}}(F_{\text{binarize}}(\mathbf{P}_i, \tau), S_{\text{min}})$\;
            $s_i \leftarrow F_{\text{DSC}}(\hat{\mathbf{Y}}_i, \mathbf{Y}_i)$\;
            $\mathcal{L}_{\text{scores}} \leftarrow \mathcal{L}_{\text{scores}} \cup \{s_i\}$\;
        }
        $\bar{s} \leftarrow F_{\text{mean}}(\mathcal{L}_{\text{scores}})$\;
        \If{$\bar{s} > \mathcal{S}_{\text{best}}$}{
            $\mathcal{S}_{\text{best}} \leftarrow \bar{s}$; $\tau_{\text{opt}} \leftarrow \tau$; $S_{\text{min,opt}} \leftarrow S_{\text{min}}$\;
        }
    }
}

\tcp{Stage 2: Final Evaluation on Test Set}
$\mathcal{R}_{\text{metrics}} \leftarrow \emptyset$\;
\ForEach{$(\mathbf{X}_j, \mathbf{Y}_j) \in \mathcal{D}_{\text{test}}$}{
    $\mathbf{P}_j \leftarrow F_{\text{inference}}(\mathbf{X}_j, \mathcal{M})$\;
    $\hat{\mathbf{Y}}_j \leftarrow F_{\text{postprocess}}(F_{\text{binarize}}(\mathbf{P}_j, \tau_{\text{opt}}), S_{\text{min,opt}})$\;
    $\mathcal{M}_j \leftarrow F_{\text{calculate\_metrics}}(\hat{\mathbf{Y}}_j, \mathbf{Y}_j)$\;
    $\mathcal{R}_{\text{metrics}} \leftarrow \mathcal{R}_{\text{metrics}} \cup \{\mathcal{M}_j\}$\;
}
\KwRet{$\mathcal{R}_{\text{metrics}}, \tau_{\text{opt}}, S_{\text{min,opt}}$}\;
\end{algorithm}

\subsection{Evaluation Metrics}
\label{subsec:evaluation_metrics}

To assess performance consistency, we report the mean and standard deviation ($\pm$ SD) for the following metrics.

The primary measure for volumetric accuracy is the Dice Similarity Coefficient (DSC), defined in \eqref{eq:dsc}:
\begin{equation}
\label{eq:dsc}
    \text{DSC}(Y, \hat{Y}) = \frac{2 \cdot |Y \cap \hat{Y}|}{|Y| + |\hat{Y}|} = \frac{2 \cdot \text{TP}}{2 \cdot \text{TP} + \text{FP} + \text{FN}}
\end{equation}
where TP, FP, and FN represent voxel-level true positives, false positives, and false negatives.

To evaluate boundary delineation while mitigating outliers, we use the 95th Percentile Hausdorff Distance (HD95). Given surface point sets $\partial Y$ and $\partial\hat{Y}$, the formulation is shown in \eqref{eq:hd95}:
\begin{equation}
\label{eq:hd95}
\begin{split}
    \text{HD95}(Y, \hat{Y}) = \max\Biggl( & \mathcal{P}_{95}\left\{ \min_{b \in \partial Y} \|a - b\| \; \forall a \in \partial\hat{Y} \right\}, \\
                                          & \mathcal{P}_{95}\left\{ \min_{a \in \partial\hat{Y}} \|b - a\| \; \forall b \in \partial Y \right\} \Biggr)
\end{split}
\end{equation}
where $a$ and $b$ denote surface points, $\mathcal{P}_{95}$ is the 95th percentile, and $\|\cdot\|$ is the Euclidean distance.

Volumetric error is quantified by the Absolute Volume Difference (AVD), calculated as in \eqref{eq:avd}:
\begin{equation}
\label{eq:avd}
    \text{$\%$AVD} = \frac{|V_{\text{gt}} - V_{\text{pred}}|}{V_{\text{gt}}} \times 100\%
\end{equation}
where $V_{\text{pred}}$ and $V_{\text{gt}}$ are the predicted and ground truth volumes.

Finally, individual lesion detection is evaluated using 3D connected-component analysis. Let $N_{TP_L}$, $N_{FN_L}$, and $N_{FP_L}$ denote the counts of true positive, false negative, and false positive lesions, respectively. We report Lesion Recall to measure the fraction of detected lesions is given by \eqref{eq:recall}:
\begin{equation}
\label{eq:recall}
    \text{Recall}_{\text{L}} = \frac{N_{TP_L}}{N_{TP_L} + N_{FN_L}}
\end{equation}
and the Lesion F1-Score for a balanced detection measure is calculated as \eqref{eq:f1}:
\begin{equation}
\label{eq:f1}
    \text{F1}_{\text{L}} = \frac{2 \cdot N_{TP_L}}{2 \cdot N_{TP_L} + N_{FP_L} + N_{FN_L}}
\end{equation}
Higher values for DSC, Recall, and F1 denote better performance, while lower values for HD95 and AVD indicate superior accuracy.

\subsection{Quantitative Analysis of Ablation Studies}
\label{subsec:ablation_studies}

In order to analyze the impact of the proposed components of the framework, a comprehensive evaluation was performed on the WMH dataset on three aspects. We conducted ablation studies on the training pipeline and network architecture, alongside a study validating the 2D design rationale. The quantitative findings from these analyses have been discussed in Tables \ref{tab:pipeline_ablation}, \ref{tab:architectural_ablation}, and \ref{tab:2d_vs_3d_comparison}. Fig.~\ref{fig:performance_boxplot} provides a better understanding in terms of performance consistency for the pipeline and architecture ablation.

\begin{table*}[t]
\centering
\begin{threeparttable}
\caption{Ablation Study of Training Pipeline Components on WMH Dataset}
\label{tab:pipeline_ablation}
\scriptsize
\begin{tabular*}{\textwidth}{@{\extracolsep{\fill}}lccccc@{}}
\toprule
\textbf{Training Protocol} & \textbf{DSC $\uparrow$} & \textbf{Lesion F1 $\uparrow$} & \textbf{Recall $\uparrow$} & \textbf{HD95 $\downarrow$} & \textbf{AVD $\downarrow$} \\
\midrule
1. Minimal Pipeline & 0.765 ± 0.168 & 0.66 ± 0.211 & 0.68 ± 0.224 & 7.95 ± 5.81 & 31.54 ± 25.98 \\
2. + Robust Preprocessing & 0.791 ± 0.145 & 0.71 ± 0.185 & 0.72 ± 0.198 & 6.23 ± 4.95 & 25.88 ± 22.43 \\
3. + Difficulty-Aware Sampling & 0.816 ± 0.117 & 0.804 ± 0.136 & 0.83 ± 0.125 & 5.98 ± 4.52 & 18.72 ± 16.91 \\
4. + Composite Loss(Complete) & \textbf{0.831} ± 0.093 & \textbf{0.816} ± 0.105 & \textbf{0.84} ± 0.107 & \textbf{3.03} ± 2.25 & \textbf{13.46} ± 12.44 \\
\bottomrule
\end{tabular*}
\begin{tablenotes}
\scriptsize
\item \textit{Note}: All experiments conducted with fixed SYNAPSE-Net architecture. Performance metrics demonstrate incremental improvements with each added pipeline component.
\end{tablenotes}
\end{threeparttable}
\end{table*}

\begin{table*}[t]
\centering
\begin{threeparttable}
\caption{Architectural Ablation Study on WMH Dataset}
\label{tab:architectural_ablation}
\scriptsize
\begin{tabular*}{\textwidth}{@{\extracolsep{\fill}}lccccc@{}}
\toprule
\textbf{Model Configuration} & \textbf{DSC $\uparrow$} & \textbf{Lesion F1 $\uparrow$} & \textbf{Lesion Recall $\uparrow$} & \textbf{HD95 $\downarrow$} & \textbf{AVD $\downarrow$} \\
\midrule
1. Baseline (Standard UNet++) & 0.772 ± 0.145 & 0.75 ± 0.170 & 0.80 ± 0.161 & 7.15 ± 5.88 & 23.41 ± 20.15 \\
2. + Advanced Bottleneck & 0.819 ± 0.112 & 0.796 ± 0.128 & 0.82 ± 0.130 & 4.88 ± 3.91 & 16.85 ± 15.02 \\
3. Full SYNAPSE-Net & \textbf{0.831} ± 0.093 & \textbf{0.816} ± 0.105 & \textbf{0.84} ± 0.107 & \textbf{3.03} ± 2.25 & \textbf{13.46} ± 12.44 \\
\bottomrule
\end{tabular*}
\begin{tablenotes}
\scriptsize
\item \textit{Note}: All models trained using the complete, optimized training pipeline. 
\end{tablenotes}
\end{threeparttable}
\end{table*}

\subsubsection{Ablation Study of the Training Pipeline}
This work experimentally evaluates the effect of our training protocols while keeping the structure of the SYNAPSE-Net fixed. The results are discussed in Table~\ref{tab:pipeline_ablation}. The baseline model (1) minimal pipeline using standard  min-max normalization combined with an ordinary BCE+Dice loss function, established the initial performance. However, this resulted in high variability, accompanied by a relatively higher boundary error, which is suboptimal for edge detection. We further tested the approach with (2) robust preprocessing, where Z-score normalization was utilized. All the metrics showed improvement, while a reduction was also seen in the standard deviation of the Dice Similarity Coefficient (DSC). We then tested our approach by employing the (3) difficulty-aware sampling strategy. As can be seen from the Table~\ref{tab:pipeline_ablation}, this strategy resulted in a significant improvement in lesion recall, which rose from 0.72 to 0.83, thus serving as direct evidence of the effectiveness of the approach for addressing the severe imbalance between the background and lesion classes. Finally, the (4) full framework utilized the composite loss, which is the combination of Focal-Tversky loss and Boundary loss. This setup resulted in the highest performance, while notable improvements were achieved for boundary metrics, HD95, and \%AVD. These improvements highlight the effectiveness of Boundary loss for improving the accuracy of the shape of the lesion and Focal-Tversky loss for improving the accuracy of the lesion regions.

\subsubsection{Ablation Study of Architectural Components}

\begin{figure}[!b]
    \centering
    \includegraphics[width=\linewidth]{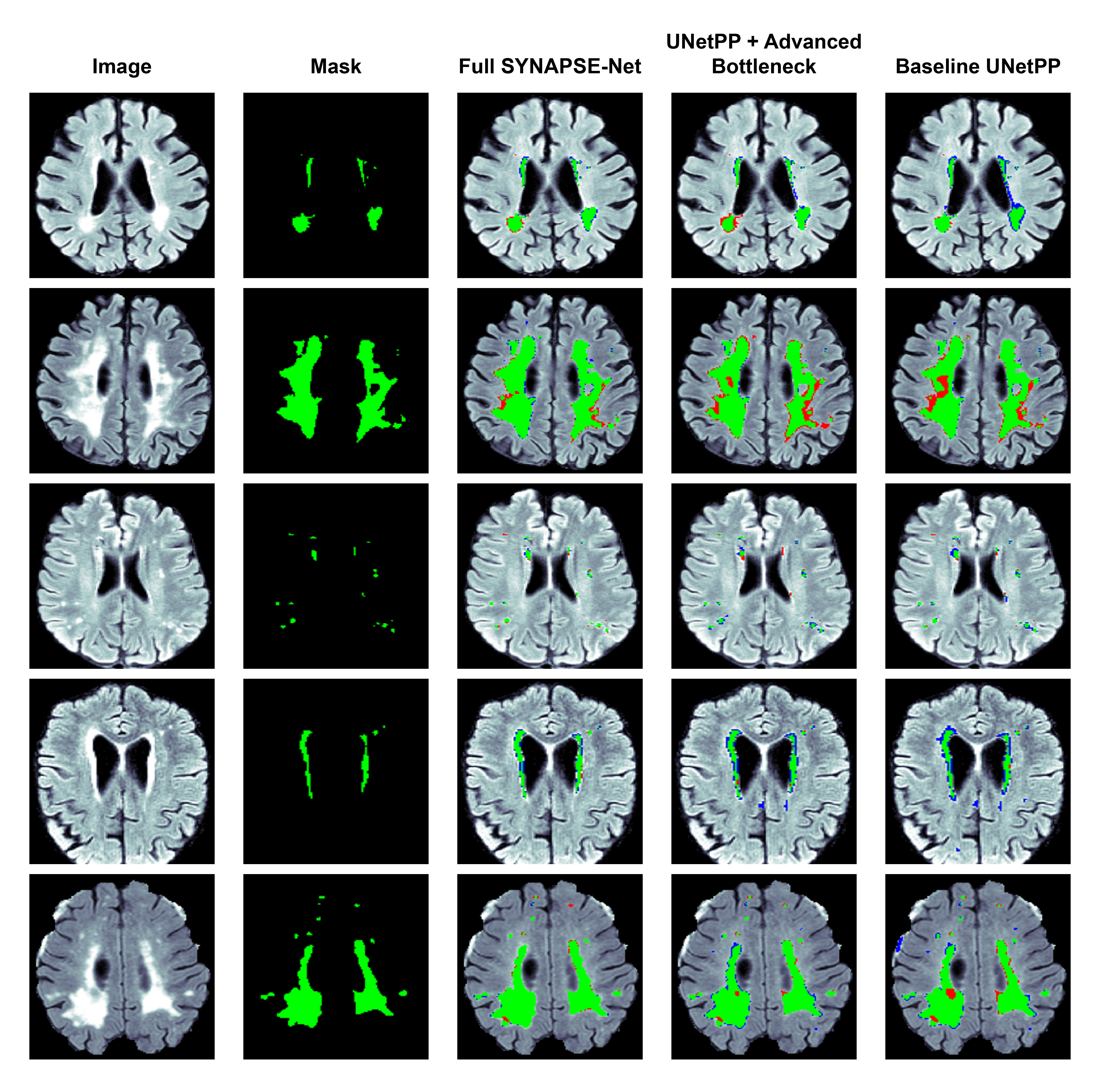} 
    \caption{Qualitative results of the architectural ablation study on the WMH dataset. Segmentation overlays show true positives (green), false negatives (red), and false positives (blue).}
    \label{fig:arch_ablation_qualitative}
\end{figure}

The next study evaluated the effects of our novel architectural components, using the complete training pipeline on each setup. Quantitative results can be found in Table \ref{tab:architectural_ablation}, and qualitative examples can be observed in Figure \ref{fig:arch_ablation_qualitative}.

\begin{figure}[htbp]
    \centering
    \includegraphics[width=\linewidth]{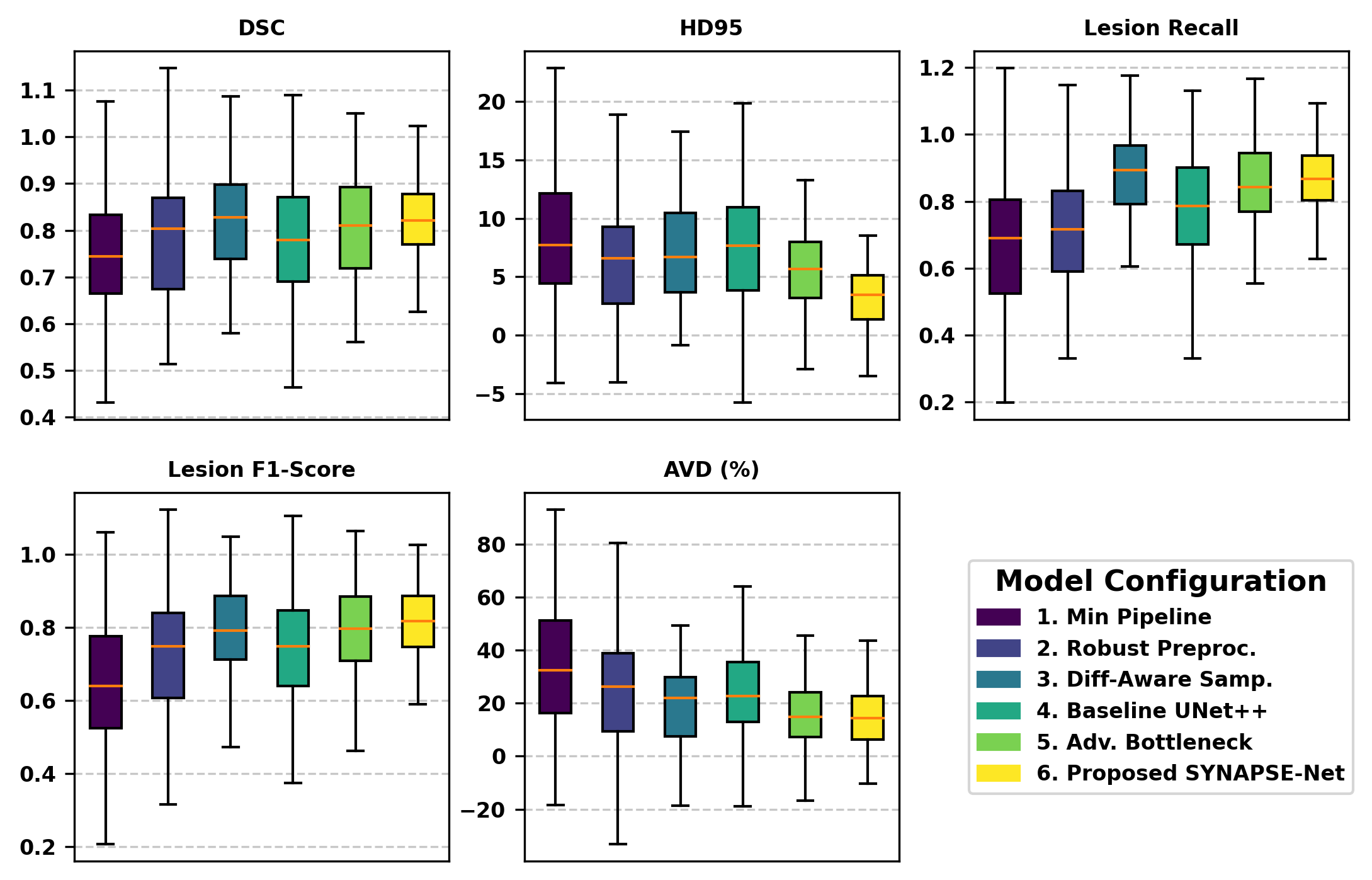}
    \caption{Boxplot visualization of key performance metrics from ablation studies.}
    \label{fig:performance_boxplot}
\end{figure}

 The baseline (1), which is the standard UNet++, had an acceptable Dice Similarity Coefficient (DSC) but poor precision, indicated by high HD95 and instability. Next, we evaluated the influence of our (2) advanced bottleneck that incorporates the multi-stream encoders, Swin Transformers, and the CMAF module. This addition yielded a significant performance boost, improving both mean DSC and HD95. The Swin Transformer's capture of global context and CMAF's effective fusion drive this performance boost. Finally, we evaluated the full SYNAPSE-Net (3), which integrates our novel hierarchical gated decoder. The results showed further improvement for the complete model, increasing the mean HD95 by 40\% compared to the previous results. This is because the proposed hierarchical gating mechanism is an essential component for maintaining the precision of the results. By implementing high-level semantic features as a means of modulating the passage of low-level spatial information, the decoder defines intricate lesion borders more accurately. This can be observed in Figure \ref{fig:arch_ablation_qualitative} since the complete model can accurately delineate the boundaries of the lesions compared to the other models. Additionally, as evident from the boxplots in Fig.~\ref{fig:performance_boxplot}, the standard deviation of all metrics reached their lowest points, demonstrating that the complete architecture is the most robust and reliable.

\begin{table*}[htbp]
\centering
\begin{threeparttable}
\caption{Quantitative Comparison Between 2D and 3D SYNAPSE-Net Variants (n=30)}
\label{tab:2d_vs_3d_comparison}
\scriptsize
\begin{tabular*}{\textwidth}{@{\extracolsep{\fill}}lccccccc@{}}
\toprule
\textbf{Variant} & \textbf{DSC $\uparrow$} & \textbf{Recall $\uparrow$} & \textbf{HD95 $\downarrow$} & \textbf{Signif.} & \textbf{\makecell{Params\\(M)}} & \textbf{\makecell{Patch\\FLOPs (G)}} & \textbf{\makecell{Full\\FLOPs (G)}} \\
\midrule
\textbf{2D} & 0.607 ± 0.25 & 0.56 ± 0.28 & 31.47 ± 12.52 & $p > 0.05$ & 13.52 & 13.12 & 138.6 \\
\textbf{3D} & 0.609 ± 0.32 & 0.57 ± 0.21 & 32.26 ± 15.17 & $p > 0.05$ & 33.36 & 1,753.8 & 60,177 \\
\bottomrule
\end{tabular*}
\begin{tablenotes}
\scriptsize
\item \textit{Note}: Paired t-tests revealed no significant differences (Dice: t(29)=0.076, p=0.94; Recall: t(29)=0.427, p=0.67; HD95: t(29)=0.652, p=0.52). FLOPs reported for full 208×208 input resolution.
\end{tablenotes}
\end{threeparttable}
\end{table*}

\subsubsection{Design Rationale: 2D vs. 3D Architecture}
\label{subsec:design_rationale}

We assessed the trade-off between model complexity and performance by benchmarking 2D and 3D configurations of SYNAPSE-Net on a representative subset of 30 patients from the WMH dataset. Experiments utilized full-resolution 3D volumes with a fixed batch size of 4. Computational metrics were evaluated for the target input resolution of $208 \times 208$. Statistical significance was calculated by paired t-tests on subject-wise Dice, Recall, and HD95 values. As seen in Table~\ref{tab:2d_vs_3d_comparison}, both models had almost equal performance on every parameter, with no significant difference statistically ($p > 0.05$) between the two models. But the computational cost of the 3D model was much more, with FLOPs over 430× that of the 2D model. With equivalent accuracy but with vastly lower computational requirements, it was decided that for all main experiments, a 2D configuration would be employed ensuring scalability and deployment efficiency without compromising segmentation quality.

\subsection{Comparison with State-of-the-Art Methods}
\label{subsec:sota_comparison}

To assess the performance of our unified framework, we benchmarked the SYNAPSE-Net against a range of leading, specialized methods on each of the three datasets.

\begin{figure*}[t]
    \centering
    \includegraphics[width=\textwidth]{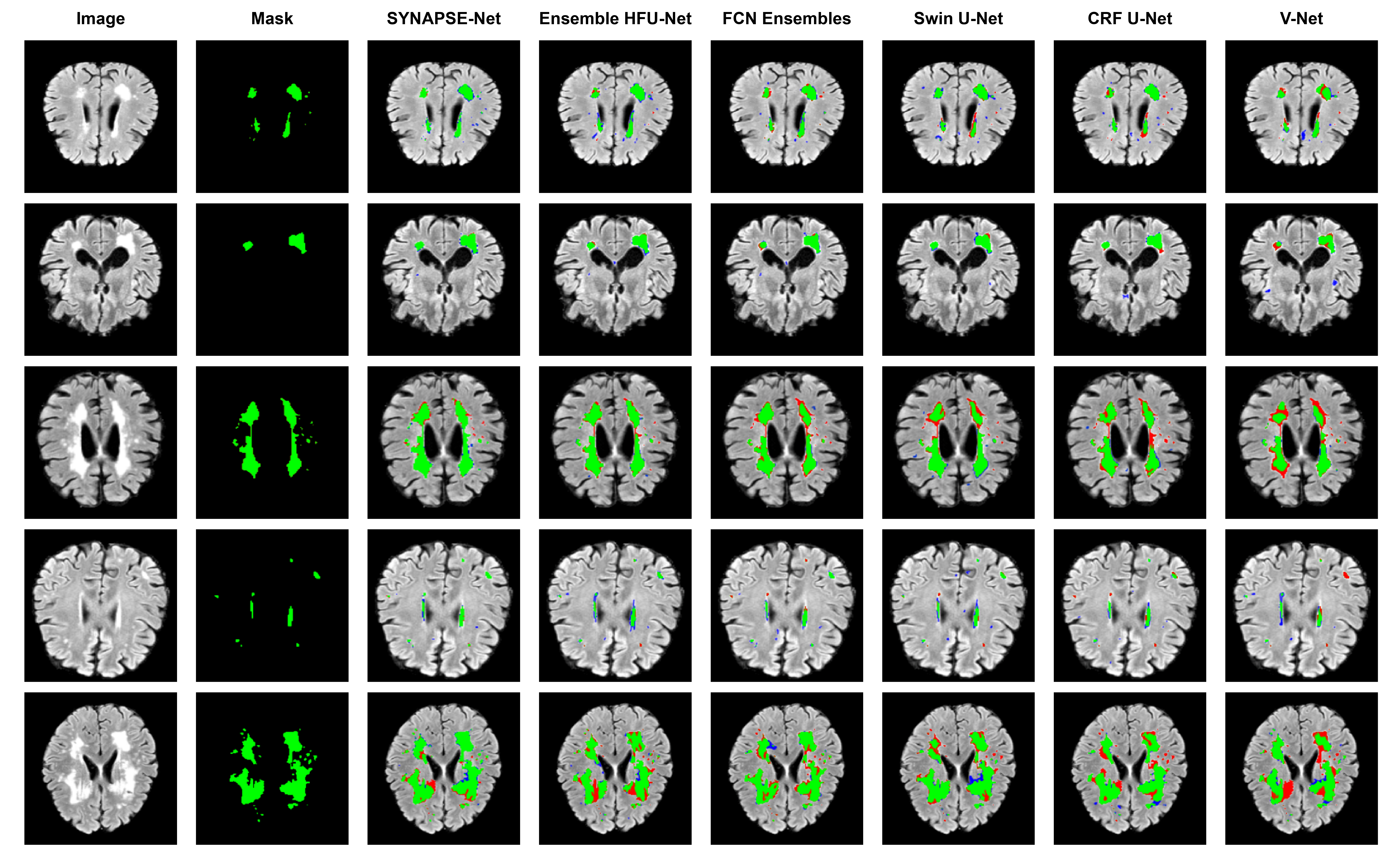}
    \caption{Qualitative comparison on the WMH test set. Green (True Positive), Red (False Negative/Missed Lesion), Blue (False Positive/Over-segmentation).}
    \label{fig:wmh_qualitative}
\end{figure*}

\begin{table*}[htbp]
\centering
\begin{threeparttable}
\caption{Performance Comparison on the MICCAI WMH 2017 Dataset}
\label{tab:miccai_wmh_reformatted}
\scriptsize
\begin{tabular*}{\textwidth}{@{\extracolsep{\fill}}lccccccc@{}}
\toprule
\textbf{Network} & \textbf{Code Avail.} & \textbf{DSC $\uparrow$} & \textbf{F1 $\uparrow$} & \textbf{Recall $\uparrow$} & \textbf{AVD $\downarrow$} & \textbf{HD95 $\downarrow$} & \textbf{Parameters} \\
\midrule
Ensemble HFU-Net \citep{Park2021} & \cmark & 0.81\tnote{*} & 0.79\tnote{*} & 0.82\tnote{*} & 18.58\tnote{*} & 5.63\tnote{*} & 8.5M \\
FCN Ensembles \citep{Li2018} & \cmark & 0.80\tnote{*} & 0.76\tnote{*} & 0.84\tnote{*} & 21.88\tnote{*} & 6.30\tnote{*} & 8.8M \\
Swin U-Net \citep{Viteri2022} & \cmark & 0.80\tnote{*} & 0.63\tnote{*} & 0.72\tnote{*} & 16.93\tnote{*} & 3.16\tnote{+} & 28.8M \\
Skip Connection U-Net \citep{Wu2019} & \cmark & 0.7836\tnote{*} & 0.7086\tnote{*} & 0.8149\tnote{*} & 28.23\tnote{*} & 7.36\tnote{*} & 10.56M \\
Attention SwinU-Net \citep{He2025} & \xmark & 0.72\tnote{*} & N/A & \textbf{0.92}\tnote{*} & 50\tnote{*} & 9.47\tnote{*} & N/A \\
WMH-DualTasker \citep{Wu2025} & \xmark & 0.602\tnote{*} & 0.677\tnote{*} & 0.758\tnote{*} & 33.8\tnote{*} & N/A & N/A \\
CRF U-Net \citep{Zhou2020} & \xmark & 0.78\tnote{*} & 0.67\tnote{*} & 0.77\tnote{*} & \textbf{10}\tnote{*} & 3.70\tnote{*} & N/A \\
V-Net \citep{Huang2023} & \xmark & 0.75\tnote{*} & 0.72\tnote{*} & 0.61\tnote{*} & 30.6\tnote{*} & 6.04\tnote{*} & N/A \\
SegAE \citep{Atlason2019} & \xmark & 0.62\tnote{*} & 0.36\tnote{*} & 0.33\tnote{*} & 44.19\tnote{*} & 24.49\tnote{*} & N/A \\
ResU-net \citep{Jin2018} & \cmark & 0.75\tnote{*} & 0.69\tnote{*} & 0.81\tnote{*} & 27.26\tnote{*} & 7.35\tnote{*} & 28.86M \\
\midrule
\textbf{SYNAPSE-Net (ours)} & & \textbf{0.831} & \textbf{0.816} & 0.84 & 13.46 & \textbf{3.03} & 13.46M \\
\bottomrule
\end{tabular*}
\begin{tablenotes}
\scriptsize
\item \textit{Note}: Bold indicates the best performance for each metric. Code Availability indicates if the method's code was publicly available for qualitative comparison. N/A indicates parameter count not available.
\item Statistical significance of differences from our model, obtained using paired two-tailed \textit{t}-tests at a 95\% confidence level, is denoted as follows: \tnote{(*)} for $p < 0.05$ (statistically significant) and \tnote{(+)} for $p > 0.05$ (not significant).
\end{tablenotes}
\end{threeparttable}
\end{table*}

\subsubsection{WMH Segmentation Performance}
We evaluated SYNAPSE-Net on MICCAI 2017 WMH data for comparison with ten contemporary models. In Table~\ref{tab:miccai_wmh_reformatted}, the quantitative results demonstrate that our proposed framework achieves a superior and more balanced performance profile. SYNAPSE-Net achieved the highest DSC of 0.831 and a lesion-wise F1 score of 0.816, indicating a better balance of precision and recall at both voxel and object levels. The DSC of our model was significantly higher than that of the SOTA ensemble methods such as the HFU-Net ensemble \citep{Park2021} and the FCN Ensembles \citep{Li2018}. Though CRF U-Net \citep{Zhou2020} had the highest \%AVD, our model achieved \%AVD of 13.46 which is highly competitive. The most significant contribution of our framework is highlighted by its exceptional performance on boundary awareness. As detailed in Table~\ref{tab:miccai_wmh_reformatted}, our model achieved the best HD95 score of 3.03. It outperforms the majority of methods (e.g., by more than 50\% compared to 7.35 of ResU-Net \citep{Jin2018}) and performs comparably to 3.16 of Swin U-Net \citep{Viteri2022}. This quantitative superiority in boundary awareness is visually substantiated in Fig.~\ref{fig:wmh_qualitative}. In the cases of large, confluent lesions (row 3 and 5), methods such as ResU-Net and SCU-Net \citep{Wu2019} display under-segmentation and poorly estimated edges, as evident by their larger HD95 values. 

\begin{figure*}[t]
    \centering
    \includegraphics[width=\textwidth]{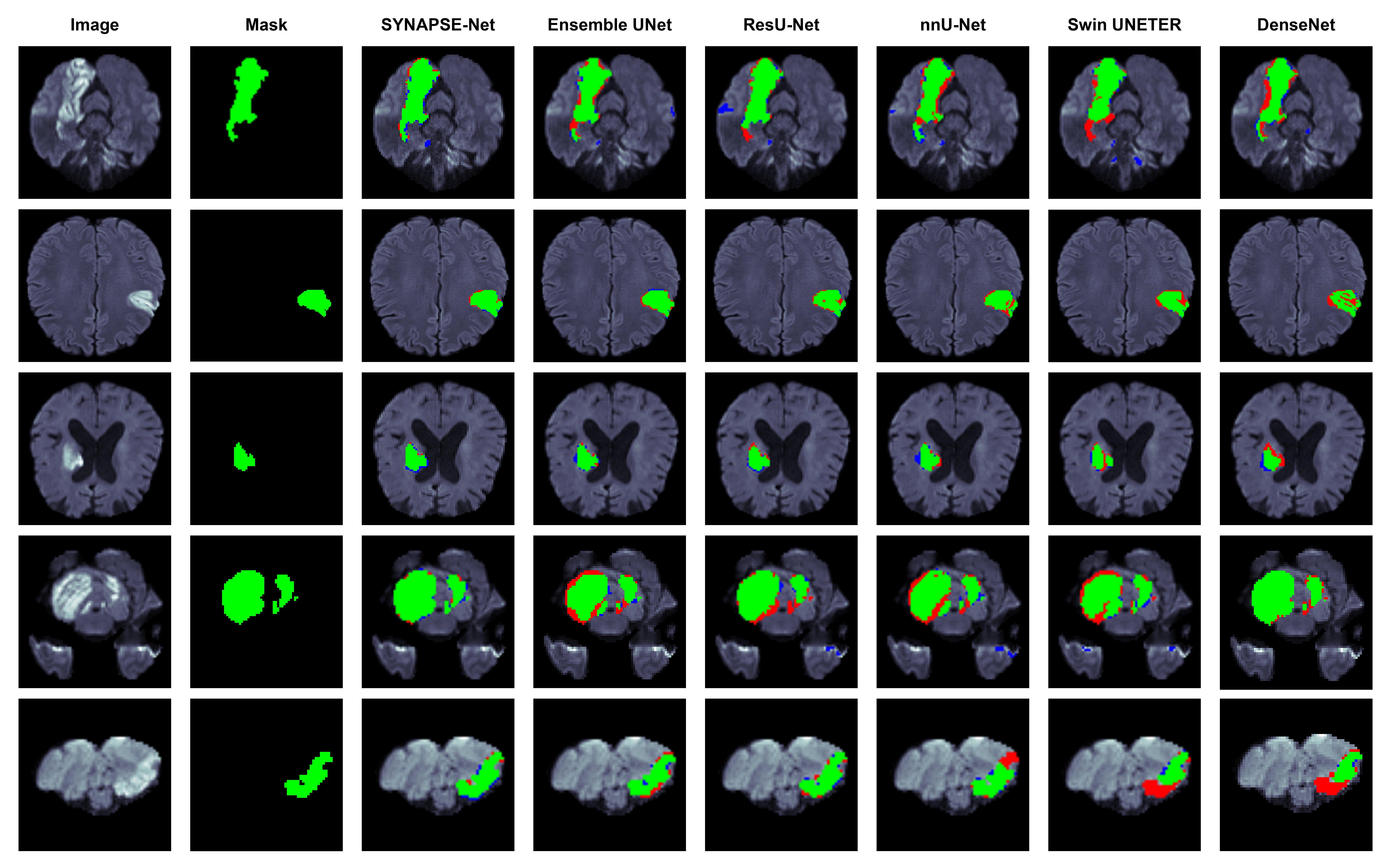}
    \caption{Qualitative comparison on the ISLES 2022 dataset. Green (True Positive), Red (False Negative), Blue (False Positive).}
    \label{fig:isles_qualitative}
\end{figure*}

\begin{table*}[t]
\centering
\begin{threeparttable}
\scriptsize
\caption{Performance Comparison on the ISLES 2022 Dataset}
\label{tab:isles22_comparison}
\begin{tabular*}{\textwidth}{@{\extracolsep{\fill}}lccccc@{}}
\toprule
\textbf{Network} & \textbf{DSC $\uparrow$} & \textbf{F1 $\uparrow$} & \textbf{Recall $\uparrow$} & \textbf{HD95 $\downarrow$} & \textbf{Parameters} \\
\midrule
nnU-Net & 0.6982\tnote{*} & 0.6567\tnote{*} & 0.5981\tnote{*} & 27.55\tnote{*} & 24.31 M \\
ResU-Net & 0.7207\tnote{*} & 0.7385\tnote{+} & 0.6691\tnote{*} & 15.77\tnote{*} & 28.86 M \\
UNETR & 0.6342\tnote{*} & 0.5953\tnote{*} & 0.5247\tnote{*} & 37.16\tnote{*} & 42.47 M \\
Swin UNETR & 0.6627\tnote{*} & 0.6213\tnote{*} & 0.5636\tnote{*} & 28.30\tnote{*} & 33.63 M \\
Ensemble UNet & 0.7580\tnote{+} & 0.7088\tnote{*} & \textbf{0.7212}\tnote{*} & 14.59\tnote{*} & 8.8 M \\
DenseNet & 0.6423\tnote{*} & 0.6019\tnote{*} & 0.5741\tnote{*} & 34.38\tnote{*} & 17.94 M \\
\midrule
\textbf{SYNAPSE-Net (ours)} & \textbf{0.7632} & \textbf{0.7420} & 0.7056 & \textbf{9.69} & 13.46 M \\
\bottomrule
\end{tabular*}
\begin{tablenotes}
\scriptsize
\item \textit{Note}: Bold indicates the best performance for each metric. Statistical significance of differences from our model, obtained using paired two-tailed \textit{t}-tests at a 95\% confidence level, is denoted as follows: \tnote{(*)} for $p < 0.05$ (statistically significant) and \tnote{(+)} for $p > 0.05$ (not significant).
\end{tablenotes}
\end{threeparttable}
\end{table*}

Our proposed framework demonstrates robust lesion detection with a recall of 0.84. Although Attention SwinU-Net \citep{He2025} achieved a higher recall of 0.92, this came at the cost of a lower DSC of 0.72, indicating over-segmentation. However, our SYNAPSE-Net achieves this high recall while maintaining the highest DSC, which proves the ability of our model to find the vast majority of lesions without sacrificing segmentation accuracy. Furthermore, SYNAPSE-Net delivers this state-of-the-art performance with just 13.46M parameters, making it more efficient than other high-performing models such as Swin U-Net (28.8M). It is visual in rows of small lesions (rows 1 and 4 of Fig.~\ref{fig:wmh_qualitative}), where our system identifies almost all small satellite lesions, in contrast to those of competing methods such as Ensemble HFU-Net and FCN Ensembles. Such detection robustness and higher boundary detection competence in Table~\ref{tab:miccai_wmh_reformatted} validate the performance of our method for WMH segmentation.

\subsubsection{Ischemic Stroke Segmentation Performance}
We evaluated the performance of SYNAPSE-Net on acute vascular lesions against a diverse set of strong baselines and state-of-the-art models. Table~\ref{tab:isles22_comparison} compares our approach with several established methods: nnU-Net \citep{Isensee2020}, ResU-Net \citep{Jin2018}, UNETR \citep{hatamizadeh_unetr_2021}, Swin UNETR \citep{Hatamizadeh2022}, and Ensemble UNet \citep{Li2018}.

Our evaluation demonstrates that SYNAPSE-Net delivers superior performance, attaining the highest DSC of 0.7632 and F1 score of 0.7420. Ensemble UNet maintains a competitive DSC of 0.7580 that does not differ significantly from ours, but its F1 performance (0.7088) is notably lower. In contrast, ResU-Net presents comparable F1 of 0.7385, which is statistically not significant from our model, although its DSC (0.7207) remains substantially lower. The most distinct advantage of our framework lies in boundary accuracy. Achieving an HD95 of 9.69, SYNAPSE-Net surpasses Ensemble UNet by 33.6\% and ResU-Net by 38.6\%. This substantial margin underscores the effectiveness of our hierarchical gated decoder and cross-modal fusion mechanism in preserving geometric fidelity. Furthermore, SYNAPSE-Net achieves these results with only 13.46M parameters. This demonstrates enhanced efficiency compared to larger models like ResU-Net (28.86M) and UNETR (42.47M), while outperforming the more compact Ensemble UNet (8.8M) in critical boundary-aware metrics.

This advantage in boundary accuracy is visually validated in Fig. \ref{fig:isles_qualitative}. In cases of large and complex infarcts (rows 1 and 4), comparison models such as Swin UNETR and DenseNet \citep{huang_densely_2017} exhibit under-segmentation along gyral boundaries, which correlates with their suboptimal HD95 scores. Conversely, our model generates a precise segmentation that closely follows the ground truth contours. The second and third rows illustrate smaller, subtle cortical and subcortical strokes where other models struggle. For instance, while Ensemble UNet achieves high recall, it is prone to scattered false positives. In contrast, our model yields a more coherent segmentation with minimal artifacts. The final row, representing a challenging cerebellar stroke, further highlights the robustness of our approach. Here, models like DenseNet and ResU-Net suffer from substantial inaccuracies, whereas our framework successfully delineates the lesion.

\begin{figure*}[t]
    \centering
    \includegraphics[width=\textwidth]{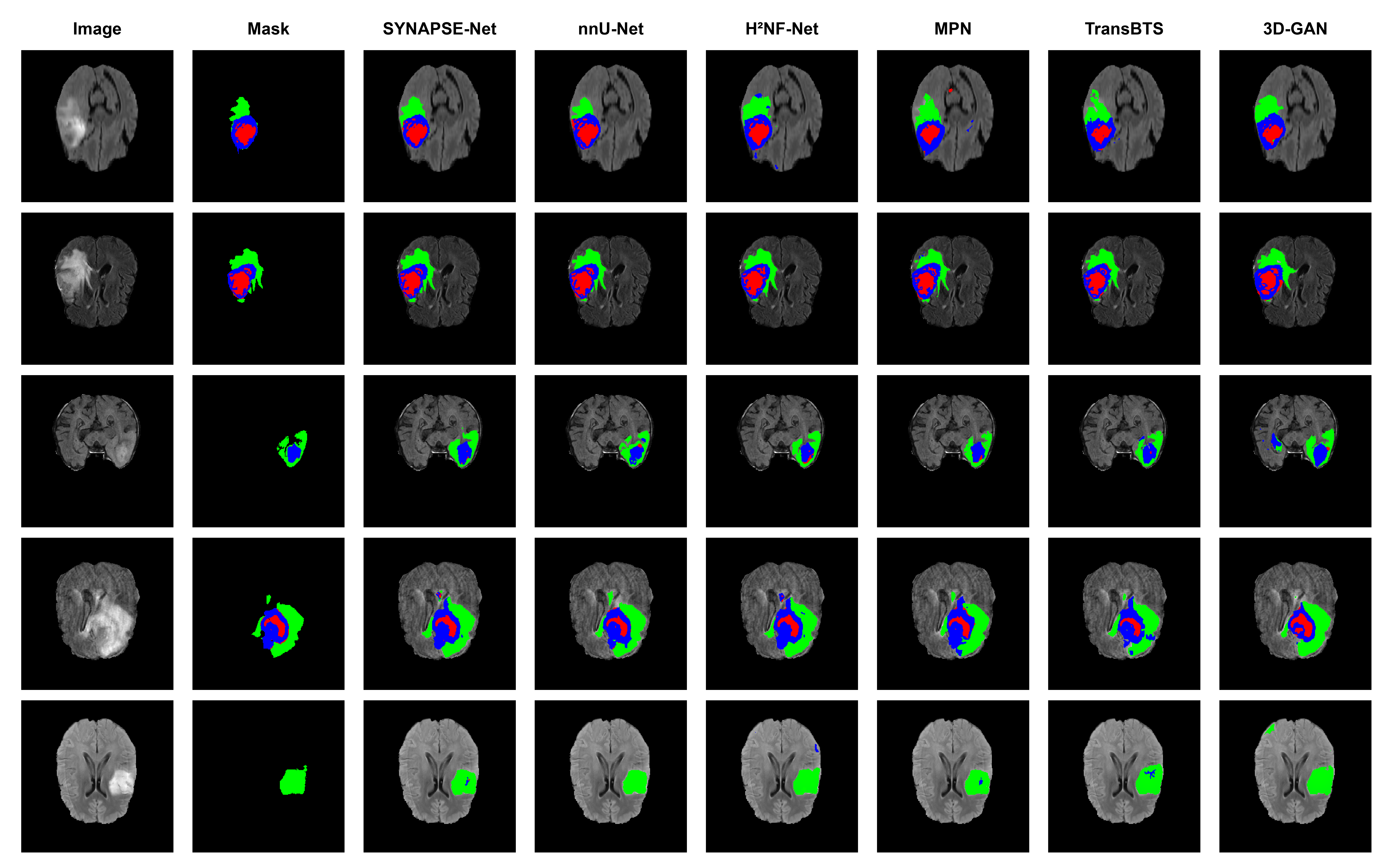} 
    \caption{Qualitative comparison on the BraTS 2020 dataset. Tumor Sub-Regions: Green (Whole Tumor), Red (Tumor Core), Blue (Enhancing Tumor).}
    \label{fig:brats_qualitative}
\end{figure*}

\begin{table*}[t]
\centering
\begin{threeparttable}
\caption{Performance Comparison on the BraTS20 Dataset}
\label{tab:brats_sota_final_correct_stats}
\scriptsize
\begin{tabularx}{\textwidth}{@{}l c *{7}{C} @{}}
\toprule
\textbf{Network} & \textbf{Code Avail.} & \multicolumn{3}{c}{\textbf{DSC $\uparrow$}} & \multicolumn{3}{c}{\textbf{HD95 $\downarrow$}} & \textbf{Parameters} \\
\cmidrule(lr){3-5} \cmidrule(lr){6-8}
 & & \textbf{WT} & \textbf{TC} & \textbf{ET} & \textbf{WT} & \textbf{TC} & \textbf{ET} & \\
\midrule
nnU-net \citep{Isensee2020} & \cmark & \textbf{0.912}$^*$ & 0.857$^*$ & 0.799$^*$ & \textbf{3.73}$^*$ & 5.64$^*$ & 26.41$^*$ & 101.88 M \\
AD-Net \citep{Peng2023} & \cmark & 0.900$^*$ & 0.800$^*$ & 0.760$^*$ & 7.22$^*$ & 15.30$^*$ & 35.20$^*$ & 5.27 M \\
MPN \citep{Wang2020} & \xmark & 0.908$^+$ & 0.856$^+$ & 0.787$^+$ & 4.71$^*$ & 5.70$^*$ & 35.01$^*$ & N/A \\
DR-Unet \citep{Colman2021} & \cmark & 0.886$^*$ & 0.669$^*$ & 0.655$^*$ & 18.39$^*$ & 53.61$^*$ & 16.19$^*$ & 76.5 M \\
3D-GAN \citep{Cirillo2020} & \xmark & 0.893$^*$ & 0.792$^*$ & 0.750$^*$ & 6.39$^*$ & 14.07$^*$ & 36.00$^*$ & N/A \\
MultiResUNet \citep{Tang2021} & \xmark & 0.893$^*$ & 0.790$^*$ & 0.703$^*$ & 4.63$^*$ & 10.07$^*$ & 34.31$^*$ & N/A \\
DLA \citep{Silva2021} & \xmark & 0.905$^+$ & 0.789$^*$ & 0.725$^*$ & 8.06$^*$ & 8.93$^*$ & 33.87$^*$ & N/A \\
SaE-Net \citep{Iantsen2021} & \cmark & 0.887$^*$ & 0.843$^*$ & \textbf{0.805}$^*$ & 4.54$^*$ & 19.59$^*$ & 15.43$^*$ & 37.59 M \\
TransBTS \citep{Wang2021} & \cmark & 0.901$^+$ & 0.817$^*$ & 0.787$^+$ & 4.96$^*$ & 9.77$^*$ & 17.95$^*$ & 32.99 M \\
\midrule
\textbf{SYNAPSE-Net (ours)} & & 0.906 & \textbf{0.865} & 0.788 & 4.32 & \textbf{4.34} & \textbf{10.94} & 41.32 M \\
\bottomrule
\end{tabularx}
\begin{tablenotes}
\scriptsize
\item \textit{Note}: Bold indicates the best performance for each metric. Code Availability indicates if the method's code was publicly available for qualitative comparison. N/A indicates parameter count not available.
\item Statistical significance of differences from our model, obtained using paired two-tailed \textit{t}-tests at a 95\% confidence level, is denoted as follows: \tnote{(*)} for $p < 0.05$ (statistically significant) and \tnote{(+)} for $p > 0.05$ (not significant).
\end{tablenotes}
\end{threeparttable}
\end{table*}

\subsubsection{Brain Tumor Segmentation Performance}
For the purpose of evaluation in the pre-surgical glioma segmentation task, SYNAPSE-Net was tested on the BraTS 2020 challenge to ensure a fair comparison between the method and the state-of-the-art CNN architectures (nnU-Net, MultiResUNet) and the Transformer architecture (TransBTS). The evaluation metrics for the Whole Tumor (WT), Tumor Core (TC), and Enhancing Tumors (ET) segments are provided in Table \ref{tab:brats_sota_final_correct_stats}.
Because the official testing data was held out by the challenge organizers for ranking purposes, we conducted our evaluation on the validation cohort and compared our findings with the results reported in \textit{Brainlesion: Glioma, Multiple Sclerosis, Stroke and Traumatic Brain Injuries, Part II}, Springer Nature Proceedings Computer Science \citep{BrainLes2018PartII}. For the WT segment, the DSC of 0.91 for our model is not significantly different from existing best models like nnU-Net \citep{Isensee2020}. The main strength of the proposed framework is its ability to capture more complex sub-regions. As seen in Table~\ref{tab:brats_sota_final_correct_stats}, our proposed model shows the best DSC of 0.87 for the TC region and the best HD95 for the TC and ET regions. These results are statistically significant, showing the improved ability to accurately delineate tumor boundaries. This performance is obtained by our model with a relatively efficient parameter count (28.62M), specifically when compared with the other models listed in the table.

The quantitative superiority is illustrated in Fig.~\ref{fig:brats_qualitative}. For instance, for the large, irregular tumor (row 1), our network provides an artifact-free segmentation, unlike TransBTS \citep{Wang2021} and DR-UNet \citep{Colman2021}, which show significant misclassifications. In row 2, our method accurately encompasses the whole hyperintensity without generating excessive false negatives compared to models like AD-Net \citep{Peng2023}. Also, on challenging cases with complex enhancing tumors (rows 3 and 4), our model demonstrates a more accurate delineation of the region with less leakage than nnU-Net and SaE-Net \citep{Iantsen2021}. Such robust and precise performance on all three tumor sub-regions validates the effectiveness of our framework in handling the heterogeneity of glioblastoma.

\section{Limitations and Future Works}
\label{sec:limitations_and_future_works}
The SYNAPSE-Net architecture has clearly shown potential for generalization over a wide range of pathologies; however, several design choices naturally limit its current scope. An important limitation of the architecture is clearly seen in the quad-modality BraTS challenge, where the hierarchical fusion process relies on the manual setting of the groups of modalities (e.g., pairing T1 with T1c), rather than learning the relationship from the data. An extension of the work could be the inclusion of an end-to-end learnable modality grouping process. Another limitation of the architecture could be seen in the post-hoc process of optimizing the thresholding, which could be improved by incorporating the uncertainty estimation process. Lastly, although our rationale for the proposed architecture clearly highlights the efficiency of the 2D process, extending the framework to fully volumetric 3D architectures remains a valuable avenue for exploration, particularly for pathologies where inter-slice contextual continuity is clinically decisive. These challenges are intended to be addressed in future iterations to further broaden the framework’s applicability.

\section{Conclusion}
\label{sec:conclusion}
This work bridges the gap between highly specialized research models and the clinical need for generalizable tools by addressing two persistent challenges: the lack of cross-pathology adaptation and the presence of performance variance. We propose SYNAPSE-Net, a general framework that leverages a strong hybrid architectural design and a training method that incorporates adaptive variance reduction. Using multi-stream encoding, a strong hybrid bottleneck, and a hierarchical gated decoder, SYNAPSE-Net goes beyond the dominance of “point solutions” in the literature, offering a single model that can be used to solve a variety of clinical problems. The efficacy of SYNAPSE-Net is shown in three different benchmark datasets. The model reaches state-of-the-art accuracy in the small, diffuse lesions of the WMH dataset (DSC 0.831) and shows improved boundary accuracy in the complex infarcts of the ISLES 2022 dataset (HD95 9.69). Moreover, it generalizes to the multi-class, quad-modality setting of BraTS 2020, correctly segmenting the key tumor sub-regions.

Ablation experiments offer empirical evidence for these results. Pipeline ablation shows that our difficulty-aware sampling and composite loss functions are not only additive, but also crucial to performance stabilization. At the same time, architectural analysis highlights the central importance of the hierarchical gated decoder, which is found to be the key factor behind geometric accuracy. In conclusion, SYNAPSE-Net shows that a general, scientifically-informed framework can provide robust and accurate segmentation of heterogeneous pathologies, and thus represents a significant step forward towards the practical application of general-purpose AI tools in real-world clinical practice.

\FloatBarrier
\section*{CRediT authorship contribution statement}

\textbf{Md. Mehedi Hassan}: Writing - original draft,Writing – review \& editing, Visualization, Validation, Software, Resources, Project administration, Methodology, Investigation, Funding acquisition, Formal analysis, Data curation, Conceptualization.

\textbf{Shafqat Alam}: Writing - original draft,Writing – review \& editing, Visualization, Validation, Software, Resources, Methodology, Investigation, Formal analysis, Data curation, Conceptualization.

\textbf{Shahriar Ahmed Seam}: Writing - review \& editing, Visualization, Validation, Software, Methodology, Investigation, Formal analysis, Data curation, Conceptualization.

\textbf{Maruf Ahmed}: Writing - review \& editing, Supervision, Resources, Project administration, Funding acquisition, Formal analysis, Investigation.

\section*{Declaration of competing interest}
The authors declare that they have no known competing financial interests or personal relationships that could have appeared to influence the work reported in this paper.

\section*{Acknowledgments}
This work was partially supported by the Basic Research Grant from Bangladesh University of Engineering and Technology (BUET).

\section*{Data Availability}
Links to the source code and the public datasets used in this study are provided in the manuscript.

\section*{Appendix: Supplementary Material}
High-resolution inference results on the test sets for all evaluated datasets are provided as supplementary materials and can be accessed via the following link:  
\url{https://tinyurl.com/5btu27ud}  

\bibliographystyle{elsarticle-harv}
\bibliography{references}

\end{document}